\documentclass[10pt,journal,compsoc]{IEEEtran}

\ifCLASSOPTIONcompsoc
  \usepackage[nocompress]{cite}
\else
  \usepackage{cite}
\fi

\ifCLASSINFOpdf

\else

\fi

\usepackage{cite}
\usepackage{amsmath,amssymb,amsfonts}
\usepackage{algorithmic}
\usepackage{graphicx}
\usepackage{textcomp}
\usepackage[table]{xcolor}
\usepackage{booktabs} 
\usepackage{times,amsmath,epsfig}
\usepackage{epstopdf}
\usepackage{enumerate}
\usepackage[ruled,vlined]{algorithm2e}
\usepackage{subfigure}
\usepackage{amsfonts}
\usepackage{amssymb}
\usepackage{graphicx}
\usepackage{diagbox}
\usepackage{multirow}
\usepackage{array}
\usepackage{bm}
\usepackage{latexsym}

\begin{document}

\title{Few-Shot Semantic Relation Prediction across Heterogeneous Graphs}

\author{Pengfei~Ding, Yan Wang, Guanfeng Liu,
        and~Xiaofang~Zhou

\IEEEcompsocitemizethanks{
\IEEEcompsocthanksitem P. Ding, Y. Wang and G. Liu are with the School of Computing, Macquarie University, Sydney NSW 2109, Australia. E-mail: pengfei.ding2@students.mq.edu.au, \{guanfeng.liu, yan.wang\}@mq.edu.au.
\IEEEcompsocthanksitem X. Zhou is with the Hong Kong University of Science and Technology, Clear Water Bay, Kowloon, Hong Kong. E-mail: zxf@cse.ust.hk.}}

\IEEEtitleabstractindextext{%
\begin{abstract}
Semantic relation prediction aims to mine the implicit relationships between objects in heterogeneous graphs, which consist of different types of objects and different types of links. In real-world scenarios, new semantic relations constantly emerge and they typically appear with only a few labeled data. Since a variety of semantic relations exist in multiple heterogeneous graphs, the transferable knowledge can be mined from some existing semantic relations to help predict the new semantic relations with few labeled data. This inspires a novel problem of few-shot semantic relation prediction across heterogeneous graphs. However, the existing methods cannot solve this problem because they not only require a large number of labeled samples as input, but also focus on a single graph with a fixed heterogeneity. Targeting this novel and challenging problem, in this paper, we propose a Meta-learning based Graph neural network for Semantic relation prediction, named MetaGS. Firstly, MetaGS decomposes the graph structure between objects into multiple normalized subgraphs, then adopts a two-view graph neural network to capture local heterogeneous information and global structure information of these subgraphs. Secondly, MetaGS aggregates the information of these subgraphs with a hyper-prototypical network, which can learn from existing semantic relations and adapt to new semantic relations. Thirdly, using the well-initialized two-view graph neural network and hyper-prototypical network, MetaGS can effectively learn new semantic relations from different graphs while overcoming the limitation of few labeled data. Extensive experiments on three real-world datasets have demonstrated the superior performance of MetaGS over the state-of-the-art methods.


\end{abstract}

\begin{IEEEkeywords}
Graph neural networks, meta-learning, heterogeneous graphs, semantic relation prediction
\end{IEEEkeywords}}

\maketitle

\IEEEdisplaynontitleabstractindextext

\IEEEpeerreviewmaketitle

\section{Introduction}
\begin{figure}[t]
\centering
\vspace{-0.1cm}
\setlength{\abovecaptionskip}{-0.05cm}   
\setlength{\belowcaptionskip}{-1cm}   
\scalebox{0.38}{\includegraphics{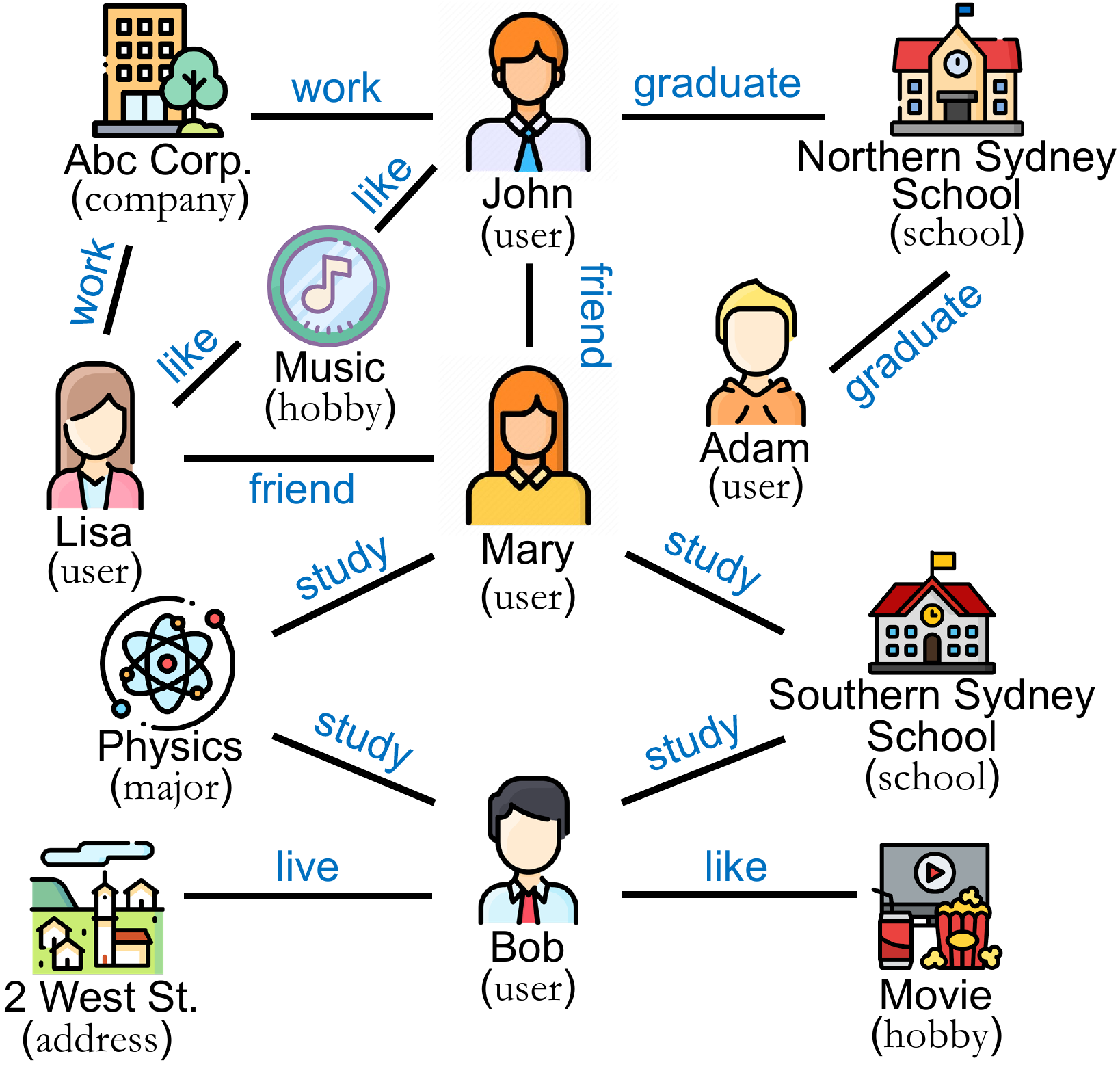}}
\caption{Toy heterogeneous graph (object type in parenthesis).}
\label{eg1}
\end{figure}

A heterogeneous graph (HG) consists of multiple types of objects (nodes) and multiple types of links (edges). Consider the heterogeneous graph in Fig. \ref{eg1} depicting a toy social network, where the objects are associated with different types (as shown in parenthesis), such as John has a type “user” and Music has a type “hobby”. These objects interact with each other via multiple typed links, such as John “works” at Abc Corp., and Mary “studies” at Southern Sydney School. In a heterogeneous graph, a variety of \emph{semantic relations} exist and can be mined \cite{liu2017semantic},\cite{zhang2022mg2vec}, which come from the combinations of various semantic meanings behind the graph structures. For example, in Fig. \ref{eg1}, the semantic relation \emph{classmate} between Mary and Bob can be mined based on the fact that they both study Physics at the same school (i.e., Southern Sydney School). In addition, John and Lisa may be \emph{friends} because they both work at the same company, Abc Corp., share a hobby, Music and have a mutual friend, Mary. Essentially, given an HG and two query objects, \emph{semantic relation prediction} aims to explore the specific semantic relations (e.g., \emph{classmate} or \emph{friend}) between the two given objects \cite{sun2011pathsim}. Studies of semantic relation prediction empower many applications. For instance, categorizing friends in social networks \cite{jung2007towards, zhou2020finding}, discovering advisors/advisees in bibliography networks \cite{xiong2017explicit}, and linking customers in e-commerce networks based on their interests  \cite{zheng2018heterogeneous}. 

In recent years, some newly emerged public events have urged a new demand for learning new semantic relations with few labeled data. For instance, when COVID is diagnosed with a few patients, the ability to accurately forecast people who are likely to be infected by these COVID patients, is extremely important to assist policymakers in rapidly controlling outbreaks. In this case, although the new semantic relation between COVID patients and susceptible people has never been learned before, researchers can mine the knowledge from similar well-studied pandemics, e.g., learn transmission rules and patterns between patients of these pandemics, to help predict people who are at high risk of being infected, with the support of a few diagnosed samples. The above discussion leads to a novel and challenging problem of how to transfer knowledge from existing semantic relations, to predict new semantic relations with only few observed data, namely, \emph{few-shot semantic relation prediction}.

In general, the studies of few-shot semantic relation prediction focus on exploiting transferable knowledge from semantic relations with a large amount of labeled data, to help learn other semantic relations with few labeled data. Therefore, such studies may also be utilised to tackle the cold-start problem and the imbalanced distribution of some particular semantic relations. In the context of this paper, we use the term “existing semantic relations” to refer to the semantic relations for which sufficient labeled data exists and can be used to learn the knowledge of these semantic relations (e.g., characteristics, formation rules or patterns of semantic relations), and use the term “new semantic relations” to refer to other semantic relations that cannot be well learned because of limited labeled data. We aim to obtain the new knowledge of these few labeled semantic relations with the support of existing semantic relations.

In real-world systems, to predict new semantic relations with few labeled data, the transferable knowledge across existing semantic relations can be mined from different scenarios as follows:

\noindent\textbf{Scenario 1. Single HG:} In this scenario, the new semantic relations to be predicted and the existing semantic relations all come from the same graph. For instance, in Fig. 2A, to predict whether “user” A and “user” B have a new semantic relation \emph{friend}, the existing semantic relations can be used to analyse the graph structure between the two users, namely, the two users have a close relationship because they are known to be \emph{schoolmates} and \emph{teammates}. Hence, the two users are quite likely to be \emph{friends}.


\noindent\textbf{Scenario 2. Multiple HGs \& single heterogeneity:} In this scenario, existing and new semantic relations come from different graphs that share a fixed \emph{heterogeneity} (i.e., a fixed set of object/link types), implying that these graphs contain different objects/links but share a common set of object/link types. For instance, in Fig. 2B, the existing semantic relations \emph{schoolmate} and \emph{friend} come from $\emph{G}_\emph{1}$ and $\emph{G}_\emph{n}$, respectively. The new semantic relation \emph{school friend} comes from another graph $\emph{G}_\emph{n+1}$ that consists of different objects, which, however, has the same object types with $\emph{G}_\emph{1}$ and $\emph{G}_\emph{n}$, such as “user”, “school” and “club”. Knowledge about these common object types' interactions can be mined from semantic relations in existing graphs (i.e., $\emph{G}_\emph{1}$-$\emph{G}_\emph{n}$), to predict the new semantic relation in $\emph{G}_\emph{n+1}$. For example, similarly to “user” C and “user” D in $\emph{G}_\emph{1}$, “user” H and “user” I attend the same school, and thus should be \emph{schoolmates}. Besides, “user” H and “user” I should be \emph{friends} because they join the same club and have a mutual friend, which is similar to the relationship between “user” E and “user” F in $\emph{G}_\emph{n}$. Therefore, “user” H and “user” I may be \emph{school friends}.



\noindent\textbf{Scenario 3. Multiple HGs \& multiple heterogeneities:} In this scenario, the existing and new semantic relations come from different graphs with different heterogeneities. For instance, in Fig. 2C, \emph{colleague} comes from $\emph{G}_\emph{1}$ that contains the objects of “employee” and “company”, and \emph{coauthor} comes from $\emph{G}_\emph{n}$ that contains the objects of “author” and “paper”. Although the two graphs contain different object types, the latent collaboration relationship can be extracted from both \emph{colleague} and \emph{coauthor}, namely, employees collaborate with each other to improve company performance, authors collaborate with each other to complete the paper. Knowledge of such collaboration relationship can be transferred to learn the new semantic relation \emph{project partner} in $\emph{G}_\emph{n+1}$, which exists between two “worker” objects that work for one “project”, even though these object types are not included in either of $\emph{G}_\emph{1}$ and $\emph{G}_\emph{n}$.


\begin{figure}[t]
\centering
\setlength{\abovecaptionskip}{-0.05cm}   
\setlength{\belowcaptionskip}{-1cm}   
\scalebox{0.203}{\includegraphics{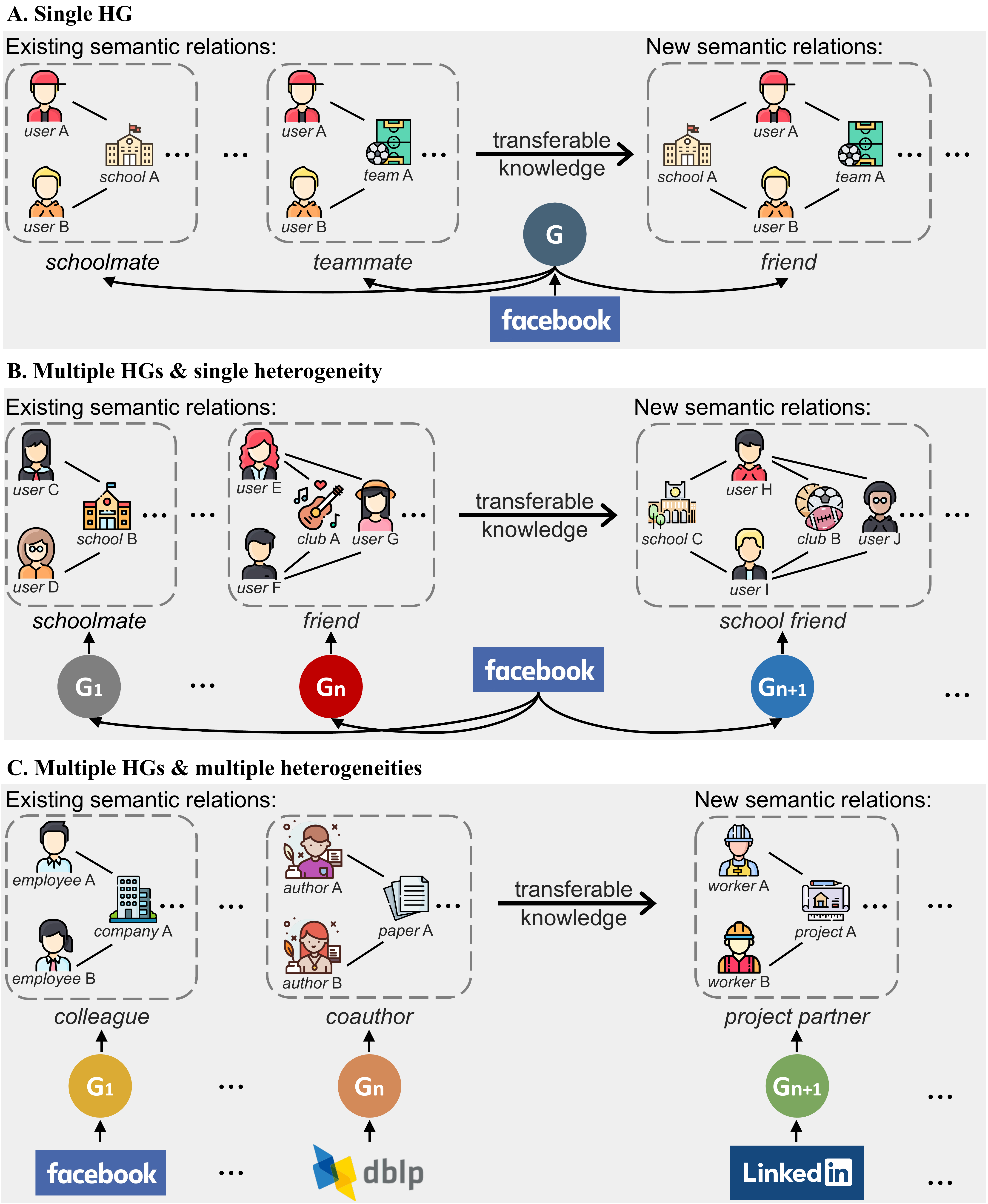}}
\caption{Examples of few-shot semantic relation prediction scenarios.}
\label{eg2}
\end{figure}






Based on the above discussions, we can see that Scenario 1 (Single HG) and Scenario 2 (Multiple HGs \& single heterogeneity) can be considered as two special cases of Scenario 3 (Multiple HGs \& multiple heterogeneities). This is because when all objects of existing and new semantic relations come from a single graph, Scenario 3 is the same as Scenario 1; and when all graphs of existing and new semantic relations share a common set of object/link types, i.e., all graphs have a fixed heterogeneity, Scenario 3 is the same as Scenario 2. In this work, we particularly focus on the most challenging Scenario 3, i.e., Multiple HGs \& multiple heterogeneities, aiming at learning transferable knowledge across semantic relations in existing graphs, to help predict new semantic relations in different graphs, with different heterogeneities and few labeled data, which is termed as the problem of \emph{few-shot semantic relation prediction across heterogeneous graphs}. We propose a novel general framework that not only solves the challenging problem, but also works for all three scenarios discussed above.

We adopt an intuitive and general idea to deal with the above-mentioned three scenarios. Specifically, we can first (1) \emph{generalize} the knowledge across existing semantic relations, then (2) \emph{transfer} the generalized knowledge to adapt to new semantic relations, and finally (3) \emph{discover} new semantic relations with the transferred knowledge and few labeled data. However, when targeting the aforementioned novel problem of few-shot semantic relation prediction across HGs, the \emph{generalize} and \emph{transfer} steps become more challenging because we need to deal with various semantic relations that come from different graphs with different heterogeneities. Based on the above discussion and proposed three steps, accordingly, we need to tackle the following three challenges of few-shot semantic relation prediction across HGs:


\noindent\textbf{CH1.} \emph{How to generalize the knowledge across existing semantic relations from multiple HGs with multiple heterogeneities?} Since the existing and new semantic relations come from graphs with different heterogeneities, each graph has its own feature space of objects/links, and the information of each semantic relation should be uniformed so that common knowledge can be extracted across existing semantic relations. However, the existing methods focus on a particular semantic relation in a particular HG \cite{fang2016semantic, liu2018interactive}, and thus cannot extract transferable knowledge across various semantic relations. 


\noindent\textbf{CH2.} \emph{How to design an efficient learning procedure to transfer knowledge from existing semantic relations to adapt to new semantic relations with few labeled data?} In the literature, some methods focus on transferring knowledge for graph problems with few labeled data, e.g., few-shot node classification \cite{lan2020node} and few-shot link prediction \cite{zhang2022few}. These methods are two-fold. On the one hand, most of these methods focus on homogeneous graphs with a single type of objects and a single type of links \cite{huang2020graph, zhou2019meta}, and thus cannot be applied to heterogeneous graphs with various object/link types. On the other hand, some methods target heterogeneous graphs \cite{zhuang2021hinfshot} and knowledge graphs (KGs) \cite{zhang2020few}. However, they focus on a particular HG/KG only, and thus cannot transfer knowledge across multiple HGs for semantic relation prediction. Therefore, it is an open challenge to design a novel framework that can transfer knowledge of semantic relations across multiple HGs.

\noindent\textbf{CH3.} \emph{How to discover effective semantic relation representations of new semantic relations from few labeled data?} To accurately predict new semantic relations with few labeled data, effectively learning and aggregating information from these few labeled samples is required. Specifically, for each labeled pair of objects, the information of the graph structure between the two objects, as well as the information of neighbor objects/links that connect the two objects, should be captured. In the literature, some existing methods leverage sequential models to learn paths connecting labeled objects for semantic relation embedding \cite{liu2017semantic, liu2018interactive}, which neglects capturing the information of the graph structure between objects. Others utilize some specifically designed patterns (e.g., “user-company-user” for \emph{colleague}) to match the graph structure between objects for semantic relation prediction \cite{sun2011pathsim},\cite{zhang2022mg2vec}. However, designing appropriate patterns requires sufficient labeled data and expertise. Therefore, with limited labeled data, it is required to develop a model that can not only capture the information of the graph structure between a pair of objects, but also learn the information of neighbor objects/links that connect the two objects.


To address the above three challenges, we propose a novel Meta-learning based Graph neural network for Semantic relation prediction, named MetaGS. Firstly, to address challenge CH1, MetaGS decomposes the graph structure between labeled objects into multiple normalized subgraphs, and then adopts a common graph neural network (GNN) to learn these subgraphs and obtain the generalized knowledge. Secondly, to address challenge CH2, MetaGS designs a novel hyper-prototypical network module to transfer knowledge across existing semantic relations, and obtain a well-initialized model that can be adapted to new semantic relations with limited data. Specifically, the hyper-prototypical network module first uses data of existing semantic relations to learn how to specify important subgraphs and aggregate information from these subgraphs, and then transfers the learned knowledge to new semantic relations. Thirdly, to address challenge CH3, MetaGS first develops a novel two-view GNN module that effectively models the information of neighbor objects/links and the graph structure of subgraphs, then adopts the well-initialized hyper-prototypical network to specify important subgraphs, and finally aggregates the information of important subgraphs to obtain the semantic relation representation for prediction. 

To the best of our knowledge, our work is the first one to propose the novel problem of few-shot semantic relation prediction and provide a general framework that works for three scenarios of the novel problem. The main contributions of this work are summarized as follows:
\begin{itemize}
    \item We propose a novel model MetaGS to generally learn transferable knowledge across semantic relations from existing heterogeneous graphs and adapt to predicting new semantic relations in different heterogeneous graphs, with different heterogeneities and few labeled data;
    \item We propose a novel two-view graph neural network module to capture information of neighbor objects/links and graph structures across multiple heterogeneous graphs, and propose a novel hyper-prototypical network module to generalize transferable knowledge for learning new semantic relations;
    \item We conduct extensive experiments to evaluate the performance of the proposed model. The experimental results illustrate the superiority of the proposed model over the state-of-the-art models in terms of evaluation metrics from three perspectives.
\end{itemize}

\section{Related Work}
Since our model is based on graph neural networks and meta-learning, in this section, in addition to reviewing the existing works of semantic relation prediction, we also introduce the existing works of graph neural networks and meta-learning on graphs.

\subsection{Semantic Relation Prediction}
Existing methods can be divided into two categories (1) pattern-based methods and (2) path-based methods. Methods in the first category investigate semantic relations by extracting predefined patterns \cite{fu2017hin2vec, sun2020updates, zhou2021butterfly}. For example, PathSim \cite{sun2011pathsim} captures semantic relations by manually selecting useful path patterns as metapaths (e.g., user-company-user). MGP \cite{fang2016semantic} mines frequent subgraph patterns as meta-graphs (e.g., user-school \& major-user). However, these methods can only mine or select patterns defined in advance, and thus require domain knowledge and sufficient samples to determine patterns when facing new semantic relations. 

To avoid manually and empirically exploring patterns, path-based methods adopt recurrent neural networks to embed paths between objects for semantic relation prediction. For instance, ProxEmbed \cite{liu2017semantic} embeds paths between objects using a recurrent neural network, and aggregates the vectors of multiple paths for prediction. IPE \cite{liu2018interactive} constructs an interactive-path structure consisting of multiple related paths, then adopts a Gated Recurrent Unit (GRU) network \cite{chung2014empirical} to embed each interactive-path for prediction. However, these methods are specialized in one semantic relation of a given HG only, and thus lack the generalization capability across HGs.



\subsection{Graph Neural Networks}
GNNs are becoming increasingly popular due to their capability of modeling graph-structured data \cite{wu2020comprehensive, hamilton2017inductive}. Typically, a GNN model iteratively updates the representation of a node by aggregating representations of its neighboring nodes and edges. More relevant to this work, some GNN methods learn the graph structure between nodes to predict the existence of links in homogeneous graphs \cite{zhang2018link,yang2021inductive} or knowledge graphs (KGs) \cite{teru2020inductive}. These methods extract the subgraph by selecting neighbors around two nodes, and then adopt a shallow GNN (1-3 layers) on the subgraph to predict whether the two nodes are physically linked. However, a semantic relation can be considered as a comprehensive representation consisting of numerous complex relationships in heterogeneous graphs, rather than a simple and intuitive relationship in homogeneous graphs or KGs.

Heterogeneous graph neural networks (HGNNs) extend GNNs to HGs to deal with various types of nodes (objects) and edges (links) \cite{wang2022survey}. Some HGNNs directly perform graph convolution on the original HGs. For example, HGAT \cite{linmei2019heterogeneous} aggregates information at the node and type levels with an attention mechanism for short-text classification. HetSANN \cite{zhang2019heterogeneous} designs an attention mechanism to aggregate multi-relational information of projected objects. Other HGNN methods use metapaths to generate graphs and apply GNN afterwards. For instance, HAN \cite{wang2019heterogeneous} applies node-level and semantic-level attention on metapath-based graphs. MAGNN \cite{fu2020magnn} applies intra-metapath and inter-metapath aggregations on metapath instances. To some extent, HGNNs can be used to embed graph structures in heterogeneous graphs for semantic relation prediction. However, these HGNNs are specialized in learning particular object/link types or metapaths in a single HG, and thus cannot be generalized to deal with multiple HGs.

\subsection{Meta-learning on Graphs} 
In recent years, increasing research attention has been devoted to few-shot problems on graphs. Especially, the episodic meta-learning paradigm \cite{thrun2012learning} has become the most popular strategy for this problem, which transfers knowledge learned from many similar few-shot learning tasks. Based on it, Meta-GNN \cite{zhou2019meta} applies MAML \cite{finn2017model} to tackle the low-resource learning problem on graph. Furthermore, RALE \cite{liu2021relative} uses GNNs to encode graph path-based hubs and capture the task-level dependency, to achieve knowledge transfer. GPN \cite{ding2020graph} adopts Prototypical Networks \cite{snell2017prototypical} to make the classification based on the distance between the node feature and the prototypes. G-META \cite{huang2020graph} uses local subgraphs with MAML to transfer information across tasks. However, these existing methods are designed for traditional tasks such as node classification and link prediction in homogeneous graphs. In contrast, in this paper we focus on semantic relation prediction in heterogeneous graphs.

Recently, HINFShot \cite{zhuang2021hinfshot} and HG-Meta \cite{zhang2022hg} have extended meta-learning paradigms to heterogeneous graphs. However, they are limited to citation networks and can only deal with a single HG with several fixed object types (i.e., “paper”, “author”, “venue” and “institution”). They design specialized functions for object/link types or metapaths in the citation network, which cannot be generalized to deal with different object/link types in other HGs. Furthermore, the two methods are designed to learn a single object embedding for classification and are unable to capture the relationship between objects in HGs. Therefore, HINFShot and HG-Meta not only cannot deal with multiple HGs with different heterogeneities, but also cannot be used to learn semantic relations between objects for prediction.



\begin{table}[]
\vspace{-0.1cm}
\setlength{\abovecaptionskip}{-0.05cm}   
\setlength{\belowcaptionskip}{-1cm} 
\caption{Notations and explanations.}
\label{tab_nota}
\centering
\begin{tabular}{cc}
\toprule[1pt]
\textbf{Notation}                      & \textbf{Explanation}                    \\ \midrule[0.5pt]
$\emph{G}$              & heterogeneous graph      \\
$\mathcal{V}$         & set of objects          \\
$\mathcal{E}$         & set of links          \\
$\phi$         & object type mapping function  \\ 
$\psi$         & link type mapping function          \\
$\mathcal{A}$         & set of object types          \\
$\mathcal{R}$         & set of link types          \\
$\mathcal{Y}$         & set of semantic relations       \\
$\mathbf{G_\emph{train}}$         & set of training graphs       \\
$\mathcal{Y}_\emph{base}$         & set of semantic relations in $\mathbf{G_\emph{train}}$   \\
$\mathbf{G_\emph{test}}$         & set of testing graphs       \\
$\mathcal{Y}_\emph{novel}$         & set of semantic relations in $\mathbf{G_\emph{test}}$   \\
$\mathbf{D} = (\mathcal{D}_\emph{train},\mathcal{D}_\emph{val},\mathcal{D}_\emph{test})$     & meta-set       \\
$\mathcal{T}_i$         & $i$-th task of meta-learning       \\
$\mathcal{T}^\emph{spt}$        & support set       \\
$\mathcal{T}^\emph{qry}$         & query set       \\
$\mathcal{S}_{u,v}$              & graph structure between $u$ and $v$       \\
$\mathcal{SG}_{u,v}$         & subgraph extracted from $\mathcal{S}_{u,v}$       \\
$\textbf{\emph{SG}}_{u,v}$         & set of subgraphs for $(u, v)$       \\
$\emph{score}(\cdot)$          & score function of a path \\
$\emph{N}_\emph{rel}$      & number of semantic relations in each task $\mathcal{T}$\\
$\emph{N}_\emph{type}$      & number of types in each subgraph $\mathcal{SG}$ \\
$\emph{N}_\emph{subg}$      & number of subgraphs in $\textbf{\emph{SG}}_{u,v}$ \\
$\mathbf{x}_i$   & raw feature vector of object $i$ \\
\bottomrule[1pt]
\end{tabular}
\end{table}

\section{Notations and Problem Formulation}
The notations used in this paper are shown in Table \ref{tab_nota}.

\noindent\textbf{Heterogeneous Graph.} A heterogeneous graph, denoted as $\emph{G} = (\mathcal{V}, \mathcal{E}, \phi, \psi)$, consists of an object set $\mathcal{V}$, a link set $\mathcal{E}$, an object type mapping function $\phi$ : $\mathcal{V} \mapsto \mathcal{A}$, and a link type mapping function $\psi: \mathcal{E} \mapsto \mathcal{R}$. $\mathcal{A}$ and $\mathcal{R}$ denote the sets of object types and link types respectively, where $|\mathcal{A}| + |\mathcal{R}| > 2$. 


\noindent\textbf{Semantic Relation Prediction.} Given a heterogeneous graph $\emph{G}$ = $(\mathcal{V}, \mathcal{E}, \phi, \psi)$, we use $\mathcal{Y}$ = $\{y_1, \ldots, y_n\}$ to denote its set of semantic relations. Semantic relation prediction aims to find a model $f_\theta: (u,v|(u,v)\in\mathcal{V}) \mapsto \{y_1, \ldots, y_n\}$ that can accurately map two objects $(u, v)$ to the semantic relations in $\mathcal{Y}$.

\noindent\textbf{Few-shot Semantic Relation Prediction across Heterogeneous Graphs.} 
Let $\mathbf{G}$ = $\{\emph{G}_1, \ldots, \emph{G}_N\}$ denote the entire set of heterogeneous graphs. Each graph $\emph{G}_i$ = $(\mathcal{V}_i, \mathcal{E}_i, \phi_i, \psi_i)$ has its own semantic relation set $\mathcal{Y}_i$, object type set $\mathcal{A}_i$ and link type set $\mathcal{R}_i$. Graphs in $\mathbf{G}$ can be further divided into two categories: a set of training graphs $\mathbf{G_\emph{train}}$ and a set of testing graphs $\mathbf{G_\emph{test}}$. The set of semantic relations that exist in $\mathbf{G_\emph{train}}$ is denoted as $\mathcal{Y}_\emph{base}$, i.e., $\mathcal{Y}_\emph{base}$ = $\{y|y\in \mathcal{Y}_i, \emph{G}_i \in \mathbf{G_\emph{train}}\}$. Similarly, the set of semantic relations that exist in $\mathbf{G_\emph{test}}$ is denoted as $\mathcal{Y}_\emph{novel}$. $\mathbf{G}$ = $\mathbf{G_\emph{train}} \cup \mathbf{G_\emph{test}}$, $\mathbf{G_\emph{train}} \cap \mathbf{G_\emph{test}}$ = $\emptyset$ and $\mathcal{Y}_\emph{base} \cap \mathcal{Y}_\emph{novel}$ = $ \emptyset$. Graphs in $\mathbf{G_\emph{train}}$ and $\mathbf{G_\emph{test}}$ have different heterogeneities, i.e., $\mathcal{A}_i \cap \mathcal{A}_j$ = $\emptyset$ and $\mathcal{R}_i \cap \mathcal{R}_j$ = $\emptyset$, where $\emph{G}_i \in \mathbf{G_\emph{train}}$ and $\emph{G}_j \in \mathbf{G_\emph{test}}$. Generally, the number of labeled samples is abundant for semantic relations in $\mathcal{Y}_\emph{base}$ while limited for semantic relations in $\mathcal{Y}_\emph{novel}$. The goal is to learn from $\mathcal{Y}_\emph{base}$ in $\mathbf{G_\emph{train}}$, adapt to $\mathcal{Y}_\emph{novel}$ in $\mathbf{G_\emph{test}}$, and predict new semantic relations in $\mathbf{G_\emph{test}}$ with the assistance of few labeled data.

\begin{figure*}[t]
\centering
\vspace{-0.15cm}  
\setlength{\abovecaptionskip}{-0.05cm}   
\setlength{\belowcaptionskip}{-3cm}   
\scalebox{0.41}{\includegraphics{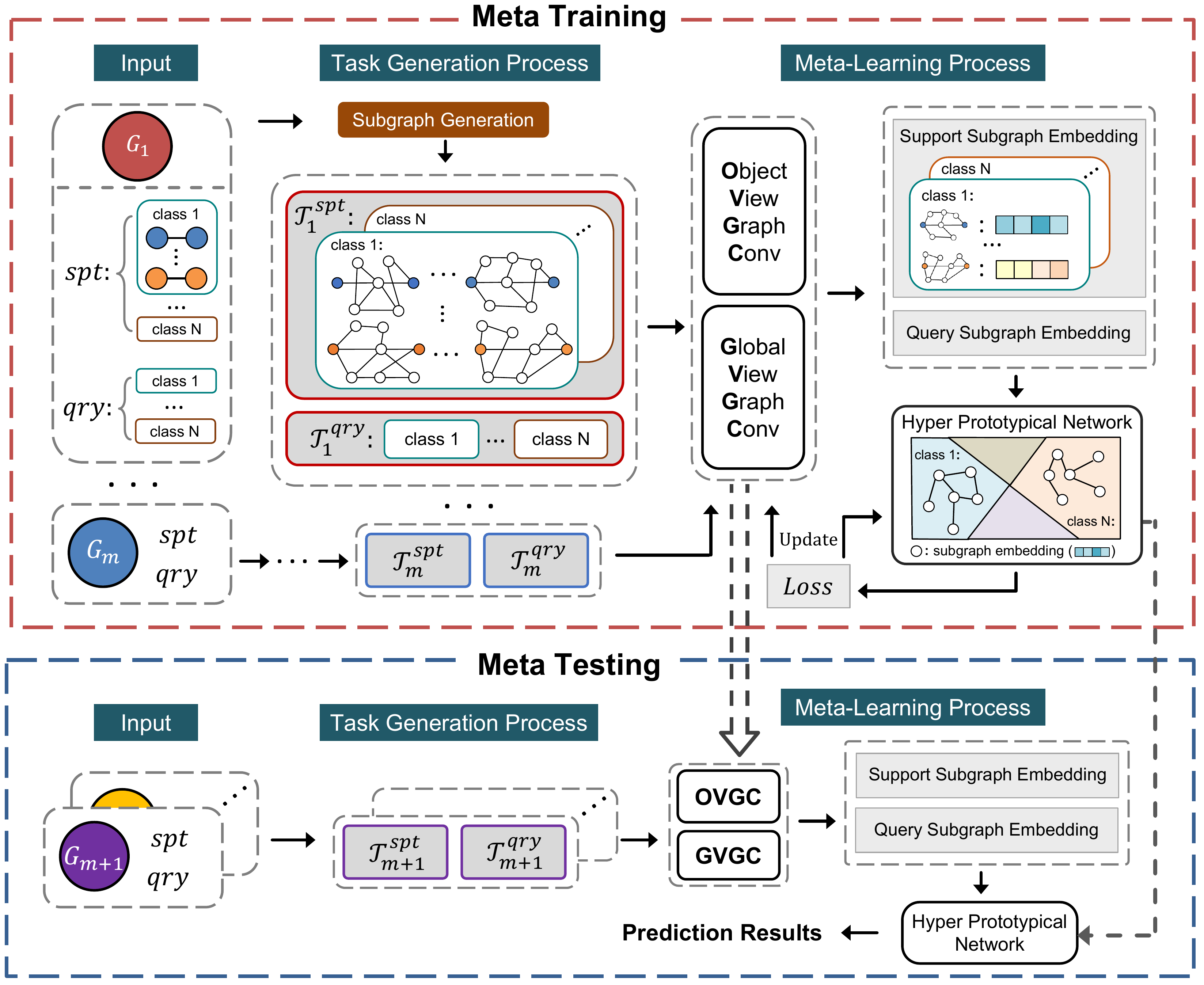}}
\caption{The overall framework of the proposed MetaGS.}
\label{model}
\end{figure*}


\noindent\textbf{Notations of Meta-learning.} In meta-learning, we have a meta-set $\mathbf{D}$ = $(\mathcal{D}_\emph{train},\mathcal{D}_\emph{val},\mathcal{D}_\emph{test})$. Each set $\mathcal{D}$ consists of many tasks $\mathcal{T}$, and each task $\mathcal{T}_i \in\mathcal{D}$ can be divided into a support set $\mathcal{T}^\emph{spt}_i$ for learning multiple semantic relations, and a query set $\mathcal{T}^\emph{qry}_i$ for evaluating these semantic relations. Specifically, $\mathcal{T}^\emph{spt}_i$ contains data of $\emph{N}_\emph{rel} \times \emph{K}_\emph{spt}$ labeled samples, where $\emph{N}_\emph{rel}$ is the number of semantic relations and $\emph{K}_\emph{spt}$ is the number of support samples for each semantic relation. Similarly, $\mathcal{T}^\emph{qry}_i$ contains data of $\emph{N}_\emph{rel} \times \emph{K}_\emph{qry}$ labeled samples and $\emph{K}_\emph{qry}$ is the number of query samples for each semantic relation. During meta-training, for each task $\mathcal{T}_i \in \mathcal{D}_\emph{train}$, the model first learns from $\mathcal{T}^\emph{spt}_i$ and then evaluates on $\mathcal{T}^\emph{qry}_i$ to see how well the model performs on the task. The goal of meta-learning is to obtain initialized parameters or models learned during meta-training, which can adapt to unseen tasks in $\mathcal{D}_\emph{test}$. Hyperparameters are tuned via $\mathcal{D}_\emph{val}$.

\section{The Proposed Model}
In this section, we propose a Meta-learning based Graph neural network for Semantic relation prediction, named MetaGS. Fig. \ref{model} presents the architecture of MetaGS.

\subsection{Framework Overview}
MetaGS follows the “decomposition-aggregation” scheme to learn semantic relations. The idea is inspired from most HGNNs that encode objects in an HG by aggregating heterogeneous information around these objects \cite{zhang2019heterogeneous, linmei2019heterogeneous}. Specifically, these methods firstly \emph{decompose} an HG into several homogeneous graphs based on different object/link types or metapaths, and thus each decomposed graph contains a specific aspect of heterogeneous information in the HG. Then, several GNNs are adopted to learn these graphs separately. Finally, the complex heterogeneous information in the HG can be obtained by \emph{aggregating} outputs of these GNNs.

In this work, we generalize this “decomposition-aggregation” idea for learning semantic relations in HGs. Specifically, we consider that a complex semantic relation between objects can be firstly decomposed into multiple \emph{basic-semantics} (i.e., simple relationships that convey intuitive semantics between two objects), then non-linearly aggregated by information of these basic-semantics. For example, in Fig. \ref{eg2}B, the semantic relation \emph{school friend} between users H and I can be determined by three basic-semantics: (1) they study at the same school; (2) they join the same club; (3) they have a mutual friend. The three basic-semantics are combined in parallel to form the \emph{school friend} semantic relation in $\emph{G}_\emph{n+1}$. Meanwhile, some basic-semantics and their non-linear combination modes may exist in a variety of semantic relations that come from multiple HGs. For example, the basic-semantics (2) and (3) mentioned above, as well as their parallel combination mode, can also be found in the semantic relation \emph{friend} of another graph $\emph{G}_\emph{n}$. Therefore, in this work, we aim to learn basic-semantics and their combination modes, which can help extract generalized knowledge across semantic relations. To this end, we decompose the graph structure between objects into multiple normalized subgraphs, and ensure that these subgraphs contain the information of basic-semantics. Then, we aim to learn semantic relations by non-linearly aggregating the information of these subgraphs. 

As shown in Fig. \ref{model}, MetaGS consists of two main processes: \emph{Task generation} and \emph{Meta-learning}. In the task generation process of MetaGS, we first adopt the subgraph extraction module to construct the subgraphs of labeled object pairs, then generate tasks for meta-training and meta-testing. In the meta-learning process of MetaGS, we first design a two-view GNN to model subgraphs' local heterogeneous information and global graph structure information, and then we propose a hyper-prototypical network to learn representations of semantic relations by aggregating information of these subgraphs. Finally, we adopt meta-learning based optimization to transfer the knowledge across existing semantic relations for learning new semantic relations with few labeled data.

\subsection{Task Generation Process}
The task generation process focuses on the “decomposition” part, which aims to decompose the graph structure between objects into multiple subgraphs for basic-semantic learning. Firstly, a subgraph extraction module is designed to generate subgraphs for each labeled object pair. Then, based on the subgraphs of labeled samples, a task generator module is devised to generate tasks for meta-training and meta-testing.





\subsubsection{Subgraph Extraction Module}
\label{sub_extract}
Given a heterogeneous graph $\emph{G}$ and a pair of query objects $(u, v)$, this module first extracts the graph structure between $u$ and $v$ (denoted as $\mathcal{S}_{u,v}$) in $\emph{G}$, then splits $\mathcal{S}_{u,v}$ into multiple subgraphs (denoted as $\mathcal{SG}_{u,v}$). In order to ensure (1) the graph structure $\mathcal{S}_{u,v}$ contains all possible semantic relations between $u$ and $v$ in $\emph{G}$, (2) each subgraph $\mathcal{SG}_{u,v}$ contains meaningful but not complicated semantics (i.e., basic-semantics), (3) subgraphs $\mathcal{SG}$ from HGs with different heterogeneities have unified heterogeneous information, and thus can be modeled by a general framework to capture transferable knowledge across semantic relations, MetaGS extracts subgraphs based on the following steps:

\noindent\textbf{Graph Structure Extraction.} 
MetaGS first extracts the graph structure $\mathcal{S}_{u,v}$ to ensure all possible semantic relations between $u$ and $v$ can be mined. A simple strategy to construct $\mathcal{S}_{u,v}$ is to use neighbors of $u$ and $v$ within a few hops (typically 1-3 hops). This extraction strategy is commonly used in existing works that model the graph structure between objects for link prediction \cite{zhang2018link, huang2020graph}, but it is not appropriate for semantic relation prediction. This is because on the one hand, the graph structure extracted by this strategy contains a large number of \emph{irrelevant} objects (i.e., objects that only connect with either $u$ or $v$), which may increase the heterogeneity of $\mathcal{S}_{u,v}$ (i.e., more object/link types) and generate noise when using $\mathcal{S}_{u,v}$ for semantic relation prediction. For example, in Fig. \ref{eg1}, Mary has a friend Lisa and Bob has a hobby Movie. Analysing the two irrelevant objects (i.e., Lisa and Movie) cannot help investigate the \emph{classmate} semantic relation between Mary and Bob. On the other hand, focusing on a limited range neighborhood around objects may overlook some semantic relations with longer distances between objects. For example, in the logistics network, the semantic relation between buyers and sellers may be connected by a large number of logistical centres.

Therefore, similar to existing methods that analyse paths between objects for semantic relation prediction \cite{liu2017semantic, liu2018interactive}, MetaGS constructs the graph structure $\mathcal{S}_{u,v}$ based on the paths connecting $u$ and $v$. However, mining all paths between query objects can be time-consuming and lead to a huge scale of $\mathcal{S}_{u,v}$. Meanwhile, the scales of various HGs may be different. To ensure the graph structures extracted from different graphs have similar scales, we propose a score function to select top-$\emph{K}_\emph{path}$ paths between $u$ and $v$, and then construct $\mathcal{S}_{u,v}$ based on these paths. Specifically, we consider two factors to rank paths that connect $u$ and $v$: (1) the richness of a path's heterogeneous information, i.e., the more diverse the object types in a path, the more complicated the semantic relations that can be mined; (2) the number of object type interactions in a path, i.e., the more links between distinct object types in a path, the more semantic meanings the path conveys. As a result, the score function of a path $\emph{pt}$ is defined as follows:
\begin{equation}\label{path_score}
\emph{score}(\emph{pt}) = \ln(|\emph{pt}|_\emph{type}) \cdot \frac{|\{(i,j):(i,j)\in \emph{pt} \wedge \phi(i)\neq\phi(j)\}|}{|\emph{pt}|},
\end{equation}
where $|\emph{pt}|_\emph{type}$ denotes the number of object types in path $pt$. The right part of E.q. (\ref{path_score}) represents the proportion of links that connect objects of distinct object types in all links of $pt$. Note that we do not consider path length when ranking paths. This is because when there are a limited number of labeled object pairs, extracting paths of varying lengths can help enrich the information of a new semantic relation. Next, $\mathcal{S}_{u,v}$ is constructed based on the set of top-$\emph{K}_\emph{path}$ paths $\mathcal{P}_{u,v}^{\emph{K}_\emph{path}}$ between $u$ and $v$, defined as $\mathcal{S}_{u,v}$ = $(\mathcal{V}_{u, v}, \mathcal{E}_{u,v}, \phi, \psi)$, where $\mathcal{E}_{u, v}$ = $\{(i, j) | (i, j) \in \emph{pt}, \emph{pt}\in \mathcal{P}_{u,v}^{\emph{K}_\emph{path}}\}$, $\mathcal{V}_{u, v}$ is the set of all objects in $\mathcal{E}_{u, v}$.

\noindent\textbf{Subgraph Generation.} 
This step aims to generate subgraphs $\mathcal{SG}_{u,v}$ from $\mathcal{S}_{u,v}$, and ensure each subgraph contains specific basic-semantics. MetaGS proposes a simple but effective generation method, i.e., each subgraph contains the information for a subset of all object types in $\mathcal{S}_{u,v}$. Specifically, for the graph structure $\mathcal{S}_{u,v}$ = $(\mathcal{V}_{u, v}, \mathcal{E}_{u,v}, \phi$ : $\mathcal{V} \mapsto \mathcal{A}, \psi$ : $\mathcal{E}$ $\mapsto$ $\mathcal{R})$, each subgraph can be denoted as $\mathcal{SG}_{u,v}$ = $(\mathcal{V}^\emph{S}, \mathcal{E}^\emph{S}, \phi$ : $\mathcal{V}^\emph{S} \mapsto \mathcal{A}^\emph{S}, \psi$ : $\mathcal{E}^\emph{S}$ $\mapsto$ $\mathcal{R}^\emph{S})$. The number of object types in each subgraph is fixed as $\emph{N}_\emph{type}$, i.e., $|\mathcal{A}^\emph{S}|$ = $\emph{N}_\emph{type}$, $\emph{N}_\emph{type}$ $<$ $|\mathcal{A}|$. For each subgraph, MetaGS randomly samples $\emph{N}_\emph{type}$ object types from $\mathcal{A}$ based on the distribution of object types, and the types of query objects are included (i.e., $\mathcal{A}^\emph{S}$ $\sim$ $\mathbf{P}(\mathcal{A}|{\mathcal{S}_{u,v}})$, $(\phi(u), \phi(v))\in\mathcal{A}^\emph{S}$). Then, $\mathcal{R}^\emph{S} \subseteq \mathcal{R}$ preserves link types that connect object types in $\mathcal{A}^\emph{S}$. $\mathcal{V}^\emph{S}$ = $\{i|i \in \mathcal{V}_{u, v}, \phi(i) \in \mathcal{A}^\emph{S}\}$ includes all objects with types in $\mathcal{A}^\emph{S}$, and $\mathcal{E}^\emph{S}$ = $\{(i, j) | (i, j) \in \mathcal{E}_{u,v}, i \in \mathcal{V}^\emph{S}\wedge j \in \mathcal{V}^\emph{S}\}$ contains corresponding links of objects in $\mathcal{V}^\emph{S}$. Note that the object type mapping function (i.e., $\phi$) in $\mathcal{SG}_{u,v}$ is the same as $\mathcal{S}_{u,v}$, because $\mathcal{V}^\emph{S}$ and $\mathcal{A}^\emph{S}$ are subsets of $\mathcal{V}$ and $\mathcal{A}$, respectively (likewise for the link type mapping function $\psi$).

\begin{figure}[t]
\centering
\setlength{\abovecaptionskip}{-0.05cm}   
\setlength{\belowcaptionskip}{-1cm} 
\scalebox{0.414}{\includegraphics{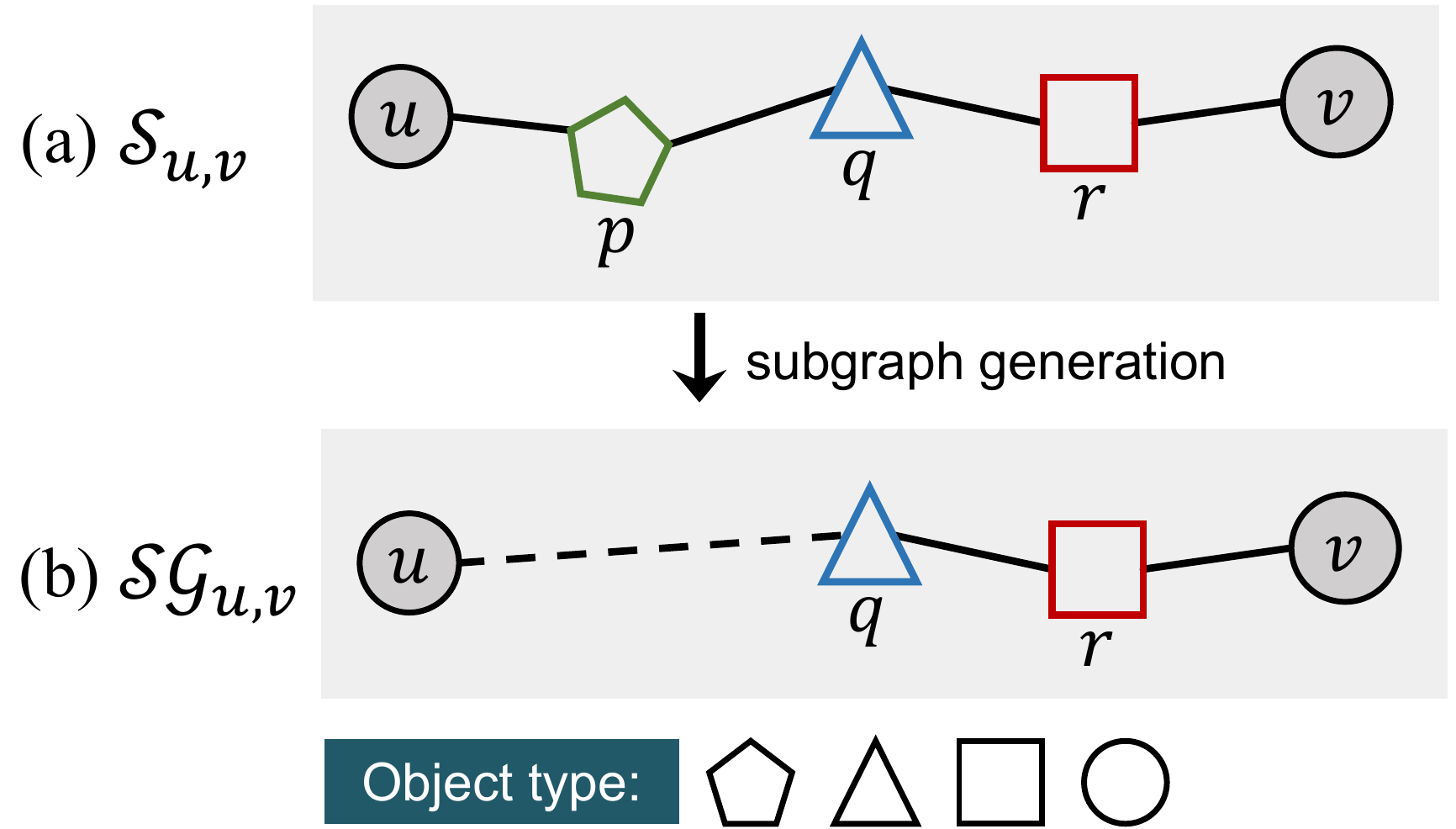}}
\caption{Situation of isolated objects in a subgraph.}
\label{eg3}
\end{figure}

For this subgraph generation strategy, there are two aspects that need to be considered: (1) The selection of $\emph{N}_\emph{type}$. With the increase of $\emph{N}_\emph{type}$, the number of object types in each subgraph increases and the heterogeneous information in each subgraph becomes sufficient. For example, when $\emph{N}_\emph{type}$ = $1$, $\mathcal{SG}_{u,v}$ becomes a homogeneous graph that consists of one object type only; when $\emph{N}_\emph{type}$ = $|\mathcal{A}|$, $\mathcal{SG}_{u,v}$ becomes $\mathcal{S}_{u,v}$. Since different HGs contain different scales and heterogeneities, it is not easy to select an appropriate value of $\emph{N}_\emph{type}$ for multiple HGs. To capture various basic-semantics, $\emph{N}_\emph{type}$ is set in the range of $[\emph{n}^\emph{min}_\emph{type}, \emph{n}^\emph{max}_\emph{type}]$, and we generate $m$ subgraphs for each setting of $\emph{N}_\emph{type}$. Thus, we can obtain $\emph{N}_\emph{subg}$ subgraphs for each pair of objects, where $\emph{N}_\emph{subg}$ = $m\times(\emph{n}^\emph{max}_\emph{type} - \emph{n}^\emph{min}_\emph{type} + 1)$, and the set of subgraphs is denoted as $\textbf{\emph{SG}}_{u,v}$ = $\{\mathcal{SG}_{u,v}^1, \ldots, \mathcal{SG}_{u,v}^{\emph{N}_\emph{subg}}\}$. (2) Isolated objects in the subgraphs. As shown in Fig. \ref{eg3}, the type of object $p$ is not selected after subgraph generation, and $q$ cannot access the query object $u$ in $\mathcal{SG}_{u,v}$. Thus, the information of object $r$ may be lost because it also cannot access $u$ via $q$. To avoid this situation, we add extra links in the subgraph generation step. Specifically, for each isolated object in $\mathcal{SG}_{u,v}$ that cannot reach both $u$ and $v$ simultaneously, the isolated object will be directly linked to the unreachable query object. For example, since $q$ can no longer access $u$ after subgraph generation, a new link $(u,q)$ is formed. With this setting, objects information in $\mathcal{S}_{u,v}$ will not be lost after generating subgraphs. Meanwhile, to minimize the number of extra links, after adding one extra link, we recheck the subgraph to filter out isolated objects that can re-access $u$ or $v$ through the newly added extra link. This process is repeated until there are no isolated objects in the subgraph.

\noindent\textbf{Object Type Mapping.} 
This step aims to unify the heterogeneous information in subgraphs. Object types in multiple subgraphs may differ after subgraph generation. To differentiate various object types and capture the generalized knowledge across semantic relations from different HGs, for each subgraph $\mathcal{SG}_{u,v}$ = $(\mathcal{V}^\emph{S}, \mathcal{E}^\emph{S}, \phi$ : $\mathcal{V}^\emph{S} \mapsto \mathcal{A}^\emph{S}, \psi$ : $\mathcal{E}^\emph{S} \mapsto \mathcal{R}^\emph{S})$, MetaGS adopts a new general type mapping function, i.e., $\phi^\emph{S}$ : $\mathcal{V}^\emph{S} \mapsto \mathcal{A}^\emph{E}$, $\mathcal{A}^\emph{E}$ {=} $\{\Phi_1, \ldots, \Phi_{\emph{N}_\emph{type}}\}$, where $\Phi_1$ is the object type in $\mathcal{A}^\emph{S}$ that is most \emph{similar} to the types of query objects (i.e., $u$ and $v$). Here we use the cosine function to calculate the similarity of two object types $(a_i, a_j)\in \mathcal{A}^\emph{S}$:
\begin{equation}
\label{type_sim}
d_\emph{type}(a_i, a_j) = \cos(\frac{\sum_{p\in \textbf{V}(a_i)}\mathbf{e}_p}{|\textbf{V}(a_i)|},\frac{\sum_{q\in \textbf{V}(a_j)}\mathbf{e}_q}{|\textbf{V}(a_j)|}),
\end{equation}
where $\textbf{V}(a_i)$=$\{p|p\in\mathcal{V}^\emph{S}, \phi(p)$ = $a_i\}$ is the set of all objects of type $a_i$ in $\mathcal{V}^\emph{S}$. Since the feature dimension spaces of different object types in an HG are typically different, existing graph embedding techniques (e.g., DeepWalk \cite{perozzi2014deepwalk}, node2vec \cite{grover2016node2vec}) can be used to encode objects in the HG into the same dimension space (i.e., $\mathbf{e}$). Then, we project object type in $\mathcal{A}^\emph{S}$ to $\mathcal{A}^\emph{E}$ based on the ranking of similarities, i.e., $\Phi_1 \emph{=} \emph{argmax}_{a_i}\{d_\emph{type}(a_i, \phi(u))+d_\emph{type}(a_i, \phi(v))|a_i\in\mathcal{A}^\emph{S}\}$ ($\Phi_{\emph{N}_\emph{type}}$ has the minimum value).

\subsubsection{Task Generator Module} 
This module aims to generate tasks with subgraphs for meta-training and meta-testing. Each task $\mathcal{T} = (\mathcal{T}^\emph{spt}, \mathcal{T}^\emph{qry})$ is designed for learning multiple semantic relations in a single graph $\emph{G}$. $\mathcal{T}^\emph{spt}$ and $\mathcal{T}^\emph{qry}$ are generated by extracting subgraphs of samples in the support set and query set, respectively. Specifically, For a graph $\emph{G}$ and its set of semantic relations $\mathcal{Y}$ = $\{y_1, \ldots, y_n\}$, $\mathcal{T}^\emph{spt}$ = $\{(\textbf{\emph{SG}}^1_1, y_1), \ldots, (\textbf{\emph{SG}}^1_{\emph{K}_\emph{spt}}, y_1), \ldots, (\textbf{\emph{SG}}^{n}_{\emph{K}_\emph{spt}}, y_n)\}$ that contains $n\times{\emph{K}_\emph{spt}}$ elements (likewise for $\mathcal{T}^\emph{qry}$), where $\emph{K}_\emph{spt}$ is the number of labeled samples for each semantic relation on support set. $\textbf{\emph{SG}}^i_j$ represents the set of subgraphs of $j$-th object pair for $i$-th semantic relation. By generating tasks on training graphs $\mathbf{G_\emph{train}}$ and testing graphs $\mathbf{G_\emph{test}}$, we can form $\mathcal{D}_\emph{train}$ and $\mathcal{D}_\emph{test}$, respectively.


\subsection{Meta-learning Process}
The meta-learning process focuses on the “aggregation” part to firstly learn subgraphs with a graph neural network, then adopt a meta-learning based framework to aggregate the information of these subgraphs for semantic relation learning. Specifically, in this process, firstly, two novel GNN based modules are designed to learn object heterogeneous information (i.e., \emph{object-view graph convolution module}) and graph structure information (i.e., \emph{graph-view graph convolution module}) of subgraphs. Then, a hyper-prototypical network is designed to learn semantic relations from subgraphs, transfer knowledge across semantic relations in meta-training, and adapt to new semantic relations with few labeled data in meta-testing.

\subsubsection{Object-View Graph Convolution Module (OVGC)} 
\label{ovgc}
OVGC aims to learn heterogeneous information around query objects (i.e.,$u$ and $v$) in the subgraph $\mathcal{SG}_{u,v}$=$(\mathcal{V}^\emph{S}, \mathcal{E}^\emph{S}, \phi^\emph{S}$ : $\mathcal{V}^\emph{S} \mapsto \mathcal{A}^\emph{E}, \psi$ : $\mathcal{E}^\emph{S} \mapsto \mathcal{R}^\emph{S})$. Specifically, {OVGC} first projects all types of object features into a common latent vector space:
\begin{equation}
\label{proj}
\mathbf{h}_i = \mathbf{W}_{\phi^\emph{S}(i)}\cdot\mathbf{x}_i,
\end{equation}
where $\mathbf{x}_i$ and $\mathbf{h}_i$ represent the original and projected feature of object $i$, respectively, $\mathbf{W}_{\phi^\emph{S}(i)}$ is the type projecting matrix for object type $\phi^\emph{S}(i)$. For query objects $(u,v)$, different types of neighbors contribute differently to their embeddings, and so do the different neighbors with the same type. So, we employ an attention mechanism here in neighbor-level and type-level to hierarchically aggregate messages from the neighbors of $(u,v)$. Specifically, we first apply neighbor-level attention to fuse query objects' neighbors with the mapped type $\Phi_k$:
\begin{equation}\label{coeff}
\mathbf{h}_u^{\Phi_k}=\emph{LeakyReLU}\left(\sum\nolimits_{i\in\mathcal{N}^{\Phi_k}_u} \alpha^{\Phi_k}_{u,i} \cdot \mathbf{h}_i\right),
\end{equation}
where $\mathcal{N}^{\Phi_k}_u$ is the neighbors with type $\Phi_k$ of $u$, $\alpha^{\Phi_k}_{u,i}$ denotes the attention value of neighbor $i$ with type $\Phi_k$ to $u$:
\begin{equation}\label{nor_fun}
\alpha^{\Phi_k}_{u,i} = \frac{\exp\left(\emph{LeakyReLU}\left(\left[\mathbf{h}_u^{\top} \| \mathbf{h}_i^{\top}\right] \cdot\mathbf{a}_{\Phi_k}\right)\right)}{\sum\limits_{j\in\mathcal{N}^{\Phi_k}_u}\exp\left(\emph{LeakyReLU}\left(\left[\mathbf{h}_u^{\top} \| \mathbf{h}_j^{\top}\right] \cdot\mathbf{a}_{\Phi_k}\right)\right)},
\end{equation}
where $\|$ denotes the concatenation operation, $\mathbf{a}_{\Phi_k}$ is the trainable attention parameter shared by the same object type $\Phi_k$.

After all type embeddings for query objects $(u, v)$ are obtained, i.e., $\{\mathbf{h}_u^{\Phi_1}, \ldots, \mathbf{h}_u^{\Phi_{\emph{N}_\emph{type}}}\}$ and $\{\mathbf{h}_v^{\Phi_1}, \ldots, \mathbf{h}_v^{\Phi_{\emph{N}_\emph{type}}}\}$, MetaGS utilizes type-level attention to fuse them together to get the final embedding $\mathbf{z}^{ov}_{u, v}$ for $\mathcal{SG}_{u, v}$ under the object view. First, we measure the weight of each object type as follows:
\begin{gather}
w_{\Phi_k} = \tanh\left({\mathbf{W}_{ov}\cdot f_\emph{agg}\left(\mathbf{h}_u^{\Phi_k}, \mathbf{h}_v^{\Phi_k}\right) + \mathbf{b}_{ov}}\right)\cdot\mathbf{a}_{ov}, \\
\beta_{\Phi_k} = \exp(w_{\Phi_k})/\sum\nolimits_{i=1}^{\emph{N}_\emph{type}}\exp(w_{\Phi_i}),  
\end{gather}
where $f_\emph{agg}(\cdot)$ is the function to aggregate the information of $u$ and $v$, e.g., $\emph{mean}(\cdot)$, $\emph{max}(\cdot)$, $\emph{sum}(\cdot)$ and concatenate. $\mathbf{W}_{ov}$ and $\mathbf{b}_{ov}$ are trainable parameters, and $\mathbf{a}_{ov}$ denotes type-level attention parameter. $\beta_{\Phi_k}$ is interpreted as the importance of type $\Phi_k$ to query objects. We weighted sum the type embeddings to get $\mathbf{z}^{ov}_{u, v}$:
\begin{equation}
    \mathbf{z}^{ov}_{u, v} = \sum\nolimits_{i=1}^{\emph{N}_\emph{type}}\beta_{\Phi_i}\cdot f_\emph{agg}\left(\mathbf{h}_u^{\Phi_i},\mathbf{h}_v^{\Phi_i}\right).
\end{equation}

OVGC focuses on the pair of query objects only and does not need to calculate embeddings and weights for other objects in subgraphs, which makes OVGC efficient. In addition, for the neighbor set in E.q. (\ref{coeff}) and (\ref{nor_fun}), i.e., $\mathcal{N}_u^{\Phi_k}$, we set $\mathcal{N}_u^{\Phi_k}$ as the set of objects that can reach the query object $u$ within $h$ hops in $\mathcal{SG}_{u,v}$ (likewise for $\mathcal{N}_v^{\Phi_k}$). Because each subgraph has a small scale, this setting of $\mathcal{N}_u^{\Phi_k}$ ensures that OVGC captures the heterogeneity information of larger surroundings with a single layer, rather than multiple layers that may cause overfitting.

\subsubsection{Graph-View Graph Convolution Module (GVGC)} 
\label{gvgc}
GVGC aggregates the structure information of all objects in the subgraph. Inspired by SEAL \cite{zhang2018link}, we propose an object labeling function to mark objects' different structural roles in subgraph $\mathcal{SG}_{u, v}$. Since SEAL focuses on homogeneous graphs, we extend its double radius vertex labeling scheme to heterogeneous graphs. Specifically, each object $i\in\mathcal{SG}_{u, v}$ is labeled with the triple $(d_\emph{srt}(i, u), d_\emph{srt}(i, v), r(i))$, where $d_\emph{srt}(i,u)$ denotes the shortest distance between $i$ and $u$ in $\mathcal{SG}_{u, v}$ (likewise for $d_\emph{srt}(i, v)$), $r(i)$ denotes the similarity ranking of type $\phi^\emph{S}(i)$ to query object types, i.e., $r(i)$ = $k$ if $i \mapsto \Phi_k$. $u$ and $v$ are uniquely labeled $(0, 1, 0)$ and $(1, 0, 0)$ so as to be identifiable. The \emph{structural feature} of $i$ is defined as:
\begin{equation}
\mathbf{h}^{st}_i = [(\emph{EnD}(d_\emph{srt}(i, u)) + \emph{EnD}(d_\emph{srt}(i, v)) \| \emph{EnD}(r(i))],
\end{equation}
where $\emph{EnD}(\cdot)$ is the one-hot encoding function, we add encodings of $d_\emph{srt}(i, u)$ and $d_\emph{srt}(i, v)$ to ensure that same type objects with the same radius get a same structural feature, e.g., objects labeled $(2, 3, 1)$ and $(3, 2, 1)$ get the same $\mathbf{h}^{st}$. Next, we generate embeddings for each object $i$ in $\mathcal{SG}_{u,v}$ with GVGC:
\begin{equation}
    \mathbf{h}^{gv}_i = \emph{GVGC}\left(\mathcal{S}_{u,v}, \left[\mathbf{h}^{st}_i\| \mathbf{h}_i\right]\right),
\end{equation}
where $\mathbf{h}_i$ is the projected feature of $i$ in E.q. (\ref{proj}). Because GVGC focuses on learning the graph structure information, GVGC is flexible with various existing GNN architectures and we adopt one GVGC to deal with all subgraphs simultaneously. Then, the final embedding $\mathbf{z}^{gv}_{u, v}$ under the graph view can be calculated by pooling embeddings of all objects in $\mathcal{SG}_{u,v}$:
\begin{equation}
    \mathbf{z}^{gv}_{u, v} = \sigma(\mathbf{W}_{gv}\cdot \emph{gv-pool}(\mathbf{h}^{gv}_{1}, \cdots, \mathbf{h}^{gv}_{|\mathcal{V}^\emph{s}|})),
\end{equation}
where $\sigma(\cdot)$ is a nonlinear activation, $|\mathcal{V}^\emph{s}|$ is the number of objects in $\mathcal{SG}_{u,v}$, $\emph{gv-pool}(\cdot)$ is the pooling function, e.g., $\emph{max-pool}(\cdot)$, $\emph{sum-pool}(\cdot)$ and $\emph{mean-pool}(\cdot)$. Finally, MetaGS concatenates the outputs of OVGC and GVGC to get the final embedding of the subgraph, i.e., $\mathbf{z}_{u, v}= \left[\mathbf{z}^{ov}_{u, v}\| \mathbf{z}^{gv}_{u, v}\right]$.


\subsubsection{Hyper-Prototypical Network Module}
\label{hyper_net}
With the learned subgraph embeddings from the two-view GNN module, next, we aim to compute the representation of each semantic relation class with few labeled samples. We follow the idea of Prototypical Networks \cite{snell2017prototypical}, which encourages samples of each class cluster around a specific prototype representation. Formally, the class prototypes can be computed by:
\begin{equation}
    \mathbf{p}_y = \textsc{Proto}\left(\{\mathbf{es}_{i}|\forall i\in \emph{Set}_y \} \right),
\end{equation}
where $\emph{Set}_y$ denotes the set of labeled samples from class $y$, $\mathbf{es}_{i}$ is the embedding of $i$-th sample. \textsc{Proto} is the prototype computation function. For instance, to solve few-shot node classification on homogeneous graphs, GPN \cite{ding2020graph} first calculates the importance of each node sample, and then weighted averages all embedded samples belonging to a class to obtain the prototype of that class:
\begin{equation}
    \mathbf{p}_y = \sum\limits_{i\in \emph{Set}_y} \gamma_i\cdot \mathbf{es}_{i},
\end{equation}
where $\gamma_i$ represents the normalized weight of each node sample. GPN \cite{ding2020graph} and other existing GNN based prototypical networks \cite{yao2020graph, tan2022graph} focus on homogeneous graphs with a single type of objects and links. These methods obtain prototypes by linearly aggregating embedded samples. However, such prototype computation functions may not be appropriate for few-shot semantic relation prediction on heterogeneous graphs, which consist of multiple types of objects and links. This is due to the fact that after encoding the samples of one semantic relation with a GNN, the embedded samples may contain heterogeneous information from multiple types of objects and links. Such heterogeneous information may have non-linear relationships to formalize the semantic relation, and thus directly adding or subtracting the embedded samples may cause our model highly noise-sensitive. Therefore, we propose a novel prototype computation function \textsc{Proto-sr} to effectively aggregate embedded samples for learning semantic relation representations. Specifically, we first extract and encode subgraphs of these samples, then treat each subgraph as a node and build a hyper-graph based on these subgraphs, and finally devise an attention-based GNN to capture non-linear relationships between subgraphs and output the prototype for a semantic relation.





\noindent\textbf{Hyper-Graph Construction.}
We construct a hyper-graph to aggregate heterogeneous information of all samples. Specifically, for one semantic relation $y$ with $\emph{K}_\emph{spt}$ labeled samples, we first extract all subgraphs of these labeled samples and form a set, i.e.,  $\textbf{\emph{SG}}^y_{spt}$ = $\{\mathcal{SG}_{1}^1, \ldots, \mathcal{SG}_{1}^{\emph{N}_\emph{subg}}, \ldots, \mathcal{SG}_{\emph{K}_\emph{spt}}^{\emph{N}_\emph{subg}}\}$, where $\mathcal{SG}^i_j$ denotes the $i$-th subgraph for $j$-th sample. We extract $\emph{N}_\emph{subg}$ subgraphs for each sample and thus $\textbf{\emph{SG}}^y_{spt}$ contains $\emph{N}_\emph{subg}\times \emph{K}_\emph{spt}$ subgraphs. Then, we put these subgraphs into the two-view GNN and obtain their embeddings, i.e., $\mathbf{Z}^y_\emph{spt}$ = $\{\mathbf{z}^1_1, \ldots, \mathbf{z}_{1}^{\emph{N}_\emph{subg}}, \ldots, \mathbf{z}_{\emph{K}_\emph{spt}}^{\emph{N}_\emph{subg}}\}$, where $\mathbf{z}^i_j$ denotes the subgraph embedding of $\mathcal{SG}^i_j$.

Next, we construct the hyper-graph $\emph{G}_\emph{hyper}$ to capture the non-linear relationships between these subgraphs, each node in $\emph{G}_\emph{hyper}$ is represented by a subgraph. Note that each $\emph{G}_\emph{hyper}$ is constructed from samples of a single semantic relation, and we aim to capture the relationship between nodes in $\emph{G}_\emph{hyper}$ to effectively learn this semantic relation. To build edges between nodes in the hyper-graph, we use the idea of \emph{heterophily} and \emph{homophily}, which is derived from \cite{rogers1970homophily}. In general, homophily refers to the fact that most connections occur among nodes with similar features, and heterophily refers to the fact that there are cases in the real world where nodes are more likely to connect when they are from different classes or have dissimilar features. For the hyper-graph construction, homophily can categorize subgraphs from all samples into multiple clusters by connecting subgraphs with similar heterogeneous information. Meanwhile, by connecting subgraphs with distinct heterogeneous information, heterophily ensures that complex non-linear combinations of multiple heterogeneous information can be formed, leading to multiple semantic relations in the hyper-graph. Specifically, we adopt two hyper-parameters to control the hyper-graph construction, i.e., \emph{homophily threshold} ($\theta_\emph{ho}$) and \emph{heterophily threshold} ($\theta_\emph{he}$). For two nodes $i, j \in \emph{G}_\emph{hyper}$, we adopt cosine function to measure the similarity between nodes in the hyper-graph, i.e., $d_\emph{hyper}(i, j)$ = $\cos(\mathbf{z}_i, \mathbf{z}_j)$, where $\mathbf{z}_i$ and $\mathbf{z}_j$ are subgraph embeddings of $i$ and $j$ respectively. Then, nodes $i$ and $j$ will be connected if the similarity $d_\emph{hyper}(i, j) > \theta_\emph{ho}$ or $d_\emph{hyper}(i, j) < \theta_\emph{he}$. It is naturally that $\theta_\emph{ho} > \theta_\emph{he}$. Moreover, by setting $\theta_\emph{ho}$ and $\theta_\emph{he}$, $\emph{G}_\emph{hyper}$ that constructed from different semantic relations will have similar uniforms, and thus can be learnt by a general framework to deal with multiple semantic relations.








\noindent\textbf{Hyper-GNN.}
After constructing the hyper-graph $\emph{G}_\emph{hyper}$, we adopt an attention-based GNN on $\emph{G}_\emph{hyper}$ to learn the prototype of the semantic relation class. Specifically, we adopt two attention modules to fuse a node's homophily neighbors (i.e., neighbors with similarities more than $\theta_\emph{ho}$) and heterophily neighbors (i.e., neighbors with similarities less than $\theta_\emph{he}$), and generate two embeddings $\mathbf{z}_i^\emph{ho}$ and $\mathbf{z}_i^\emph{he}$, respectively. Specifically, $\mathbf{z}_i^\emph{ho}$ is calculated as follows:
\begin{equation}
\label{ho1}
\mathbf{z}_i^\emph{ho-(l+1)}=\emph{LeakyReLU}\left(\sum\nolimits_{j\in\mathcal{N}^\emph{ho}_i} \alpha^\emph{ho-l}_{i,j} \cdot \mathbf{z}_j^\emph{ho-l}\right),
\end{equation}
where $\mathcal{N}^\emph{ho}_i$ is the set of homophily neighbors of $i$, $\mathbf{z}_i^\emph{ho-l}$ is the hidden state of node $i$ in the $l$-th layer and the initial state $\mathbf{z}_i^\emph{ho-0}$ = $\mathbf{z}_i$, $\alpha^\emph{ho-l}_{i,j}$ denotes the attention value of the homophily neighbor $j$ to $i$ in the $l$-th layer:
\begin{equation}
\label{ho2}
\alpha^\emph{ho-l}_{i,j} = \frac{\exp\left(\emph{LeakyReLU}\left(\left[\mathbf{z}_i^\emph{ho-l} \| \mathbf{z}_j^\emph{ho-l}\right]^\top \cdot\mathbf{a}_\emph{ho-l}\right)\right)}{\sum\limits_{u\in\mathcal{N}^\emph{ho-l}_i}\exp\left(\emph{LeakyReLU}\left(\left[\mathbf{z}_i^\emph{ho-l} \| \mathbf{z}_u^\emph{ho-l}\right] ^\top \cdot\mathbf{a}_\emph{ho-l}\right)\right)},
\end{equation}
where $\mathbf{a}_\emph{ho-l}$ is the trainable homophily attention parameter of the $l$-th layer. $\mathbf{z}_i^\emph{he}$ is computed by the same attention mechanism as $\mathbf{z}_i^\emph{ho}$, i.e., E.q. (\ref{ho1}) and (\ref{ho2}), but leverages the information of heterophily neighbors $\mathcal{N}^\emph{he}_i$, and adopts different parameters, i.e., $\mathbf{a}_\emph{he-l}$.

MetaGS adopts the outputs of $l$-th layer as the final embeddings, i.e., $\mathbf{z}_i^\emph{ho}$=$\mathbf{z}_i^\emph{ho-l}$ and $\mathbf{z}_i^\emph{he}$=$\mathbf{z}_i^\emph{he-l}$, then aggregates the nodes in the hyper-graph to represent the prototype of a semantic relation class. The new prototype computation function \textsc{Proto-sr} can be defined as follows:
\begin{equation}
\label{output}
    \mathbf{p}_y = \emph{hy-pool}\left(\left[\mathbf{z}_1^\emph{ho}\|\mathbf{z}_1^\emph{he}\right], \left[\mathbf{z}_2^\emph{ho}\|\mathbf{z}_2^\emph{he}\right], \ldots, \left[\mathbf{z}_{\emph{N}_\emph{hy}}^\emph{ho}\|\mathbf{z}_{\emph{N}_\emph{hy}}^\emph{he}\right]\right),
\end{equation}
where $\emph{N}_\emph{hy}$ is the number of nodes in the hyper-graph, $\emph{hy-pool}(\cdot)$ is the pooling function, e.g., $\emph{max-pool}(\cdot)$, $\emph{sum-pool}(\cdot)$ and $\emph{mean-pool}(\cdot)$. With this novel hyper-prototypical network, our model can capture the non-linear relationships between subgraphs to better represent the samples in both the support set and query set. Specifically, for each object pair $(u, v)$ that exists in query set or support set, the embedding of $(u, v)$ can be obtained by putting all subgraphs (i.e., $\textbf{\emph{SG}}_{u,v}$) into the hyper-prototypical network, and using the output of E.q. (\ref{output}) as the object pair embedding, denoted as $\mathbf{z}_{u,v}^\emph{hy}$.


\noindent\textbf{Loss Function.} In meta-training, after calculating the prototype for each semantic relation with samples in the support set $\mathcal{T}^\emph{spt}$, to determine the semantic relation class for a query object pair $(u, v)$ in the query set $\mathcal{T}^\emph{qry}$, the probability over each semantic relation class can be calculated based on the distance between the object pair embedding $\mathbf{z}_{u,v}^\emph{hy}$ and each prototype $\mathbf{p}_{y'}$:
\begin{equation}\label{class_vec}
p(y|u,v)=\frac{\exp \left(-d(\mathbf{z}_{u,v}^\emph{hy},\mathbf{p}_{y})\right)}{\sum_{y'} \exp \left(-d(\mathbf{z}_{u,v}^\emph{hy},\mathbf{p}_{y'})\right)},
\end{equation}
where $d(\cdot)$ is a distance metric function. Commonly, squared Euclidean distance is a simple and effective choice. Under the episodic training framework, the objective of each meta-training task is to minimize the classification loss between the predictions of the query set and the ground-truth. Specifically, the training loss can be defined as the average negative log-likelihood probability of assigning correct semantic relation labels:
\begin{equation}\label{loss_fun}
\mathcal{L} = -\frac{1}{|\mathcal{Y}|\times{\emph{K}_\emph{qry}}}\sum\limits_{i=1}^{|\mathcal{Y}|\times{\emph{K}_\emph{qry}}} \log p(y^*_i|u_i,v_i).
\end{equation}

By minimizing the above loss function, MetaGS is able to learn a generic classifier for a specific meta-training task. After training on all meta-training tasks $\mathcal{D}_\emph{train}$, its generalization performance will be evaluated on the testing tasks $\mathcal{D}_\emph{test}$.

\section{Experiments and Analysis}
We conduct extensive experiments on three real-world datasets to answer the following key questions: \textbf{Q1:} How does our MetaGS perform when compared to the-state-of-the-art models for three scenarios of few-shot semantic relation prediction (i.e., Scenario 1: Single HG, Scenario 2: Multiple HGs \& single heterogeneity, and Scenario 3: Multiple HGs \& multiple heterogeneities)? \textbf{Q2:} How do our proposed two major modules (i.e., two-view graph neural network module and hyper-prototypical network module) contribute to performance improvement? 



\subsection{Datasets} 
Social network datasets have been widely used in the studies of semantic relation prediction \cite{fang2016semantic,liu2017semantic}. Therefore, we selected three real-world datasets, i.e., Facebook, Twitter and Google+ \cite{leskovec2012learning}. Details of the datasets are shown in Table \ref{dataset}.

\begin{table}[]
\footnotesize
\centering
\setlength{\abovecaptionskip}{-0.05cm}   
\setlength{\belowcaptionskip}{-1cm}   
\caption{Statistics of datasets.}
\label{dataset}
\setlength{\tabcolsep}{0.7mm}
{
\begin{tabular}{c|c c c c c}
\toprule[1pt]
Dataset  & \#Graphs  & \#Objects    & \#Links      & Avg. types & Avg. relations \\ \midrule[0.5pt]
Facebook & 10   & 5,307   & 208,490  & 22       & 20        \\ 
Twitter  & 1,000 & 81,362 & 1,394,807 & 823      & 5         \\ 
Google+  & 132 & 107,614 & 13,673,453	 & 7      & 4         \\ \bottomrule[1pt]
\end{tabular}}
\end{table}

\begin{table*}[]
\centering
\setlength{\abovecaptionskip}{-0.05cm}   
\setlength{\belowcaptionskip}{-1cm}   
\caption{Comparison of prediction accuracy with baselines on three datasets.}
\label{exp_res}
\resizebox{182mm}{56mm}
{\setlength{\tabcolsep}{1mm}{\begin{tabular}{ccccccccccccccccccccc}
\specialrule{0.1em}{2pt}{2pt}
\multicolumn{1}{c|}{}                                                     & \multicolumn{2}{c}{\textbf{GraphSAGE}} & \multicolumn{2}{c|}{\textbf{GAT}}             & \multicolumn{2}{c}{\textbf{HAN}} & \multicolumn{2}{c|}{\textbf{MAGNN}} & \multicolumn{2}{c}{\textbf{Meta-GNN}} & \multicolumn{2}{c|}{\textbf{G-Meta}}          & \multicolumn{2}{c}{\textbf{SEAL}} & \multicolumn{2}{c|}{\textbf{SLiCE}}                      & \multicolumn{2}{c}{\textbf{MetaGS}} & \multicolumn{2}{c}{\textbf{Improvement$^2$}} \\ \specialrule{0.06em}{2pt}{2pt}
\multicolumn{1}{c|}{\multirow{-2}{*}{\textbf{}}}                                                           & \textbf{Acc}          & \textbf{F1}          & \textbf{Acc}    & \multicolumn{1}{c|}{\textbf{F1}}     & \textbf{Acc}           & \textbf{F1}           & \textbf{Acc}    & \multicolumn{1}{c|}{\textbf{F1}}     & \textbf{Acc}         & \textbf{F1}         & \textbf{Acc}    & \multicolumn{1}{c|}{\textbf{F1}}     & \textbf{Acc}        & \textbf{F1}         & \textbf{Acc}    & \multicolumn{1}{c|}{\textbf{F1}}     & \textbf{Acc}          & \textbf{F1}          & \textbf{Acc}             & \textbf{F1}            \\ \specialrule{0.05em}{2pt}{2pt}
\multicolumn{21}{c}{\cellcolor[HTML]{E5E5E5}$\mathbf{\emph{K}_\emph{spt} = 1}$}                                                                                                                                                                                                                                                                                                                                                                                               \\ \specialrule{0.06em}{2pt}{2pt}
\multicolumn{21}{c}{\textbf{Scenario 1: Single HG}}                                                                                                                                                                                                                                                                                                                                                                                              \\ \specialrule{0.06em}{2pt}{2pt}
\multicolumn{1}{c|}{\textbf{F}acebook}                                                    & 0.4085      & 0.3221      & 0.3550 & \multicolumn{1}{c|}{0.2469} & 0.4556     & 0.3134     & 0.4427 & \multicolumn{1}{c|}{0.3754*} & 0.3998        & 0.2545       & 0.4301 & \multicolumn{1}{c|}{0.2879} & {0.4652*}      & 0.2980     & 0.4548 & \multicolumn{1}{c|}{0.3551}  & \textbf{0.4975}$^1$       & \textbf{0.4553}      & 6.94\%                &  21.28\%             \\ \specialrule{0em}{1pt}{1pt}
\multicolumn{1}{c|}{\textbf{T}witter}                                                     & 0.4406      & 0.3890      & 0.4889 & \multicolumn{1}{c|}{0.3542} & N/A        & N/A        & N/A    & \multicolumn{1}{c|}{N/A} & 0.4665        & 0.3564       & 0.4208 & \multicolumn{1}{c|}{0.3196} & 0.4777      & 0.3184     & 0.5081* & \multicolumn{1}{c|}{0.4082*}     & \textbf{0.5295}       & \textbf{0.4951}      &4.21\%                 & 21.29\%              \\ \specialrule{0em}{1pt}{1pt}
\multicolumn{1}{c|}{\textbf{G}oogle+}                                                     & 0.3852      & 0.3149      & 0.3473 & \multicolumn{1}{c|}{0.2407} & 0.4336     & 0.3162     & 0.4478* & \multicolumn{1}{c|}{0.3552*} & 0.4215        & 0.2568       & 0.4314 & \multicolumn{1}{c|}{0.1918} & 0.3313      & 0.2860     & 0.4447 & \multicolumn{1}{c|}{0.3078}  & \textbf{0.4786}       & \textbf{0.4141}      &  6.88\%  & 16.58\%               \\ \specialrule{0.06em}{2pt}{2pt}
\multicolumn{21}{c}{\textbf{Scenario 2: Multiple HGs \& single heterogeneity}}                                                                                                                                                                                                                                                                                                                                                                   \\ \specialrule{0.06em}{2pt}{2pt}
\multicolumn{1}{c|}{\textbf{G}oogle+}                                                     & 0.3625      & 0.2945      & 0.3215 & \multicolumn{1}{c|}{0.2021} & 0.4268   & 0.3065        & 0.4325*    & \multicolumn{1}{c|}{0.3365*} & 0.4135        & 0.2468       & 0.4192 & \multicolumn{1}{c|}{0.1824} & 0.2914      & 0.2568     & 0.4214 & \multicolumn{1}{c|}{0.2816}     & \textbf{0.4435}       & \textbf{0.4044}      &2.54\%  & 20.18\%                \\ \specialrule{0.06em}{2pt}{2pt}
\multicolumn{21}{c}{\textbf{Scenario 3: Multiple HGs \& multiple heterogeneities}}                                                                                                                                                                                                                                                                                                                                                               \\ \specialrule{0.06em}{2pt}{2pt}
\multicolumn{1}{c|}{\textbf{F-T}} & 0.4543      & 0.3809      & 0.4294 & \multicolumn{1}{c|}{0.3247} & N/A        & N/A        & N/A    & \multicolumn{1}{c|}{N/A} & 0.5326*        & 0.3857*       & 0.5268 & \multicolumn{1}{c|}{0.3606} & 0.4439      & 0.2753     & 0.5142 & \multicolumn{1}{c|}{0.3675}     & \textbf{0.5483}       & \textbf{0.5067}      &2.95\%  & 31.37\%              \\\specialrule{0em}{1pt}{1pt}
\multicolumn{1}{c|}{\textbf{F-G}} & 0.3524      & 0.2841      & 0.3128 & \multicolumn{1}{c|}{0.1987} & N/A        & N/A        & N/A    & \multicolumn{1}{c|}{N/A} & 0.4112*        & 0.3234*       & 0.3906 & \multicolumn{1}{c|}{0.2243} & 0.2814      & 0.1999     & 0.4064 & \multicolumn{1}{c|}{0.2704}     & \textbf{0.4451}     & \textbf{0.4147}      &8.24\%  & 28.23\%               \\\specialrule{0em}{1pt}{1pt}
\multicolumn{1}{c|}{\textbf{T-F}} & 0.3823      & 0.3018      & 0.2468 & \multicolumn{1}{c|}{0.1551} & N/A        & N/A        & N/A    & \multicolumn{1}{c|}{N/A} & 0.4308        & 0.4068*       & 0.4453* & \multicolumn{1}{c|}{0.3124} & 0.3634      & 0.1755     & 0.4034 & \multicolumn{1}{c|}{0.3408}     & \textbf{0.4673}       & \textbf{0.4237}      &4.94\%  & 4.15\%               \\\specialrule{0em}{1pt}{1pt}
\multicolumn{1}{c|}{\textbf{T-G}}  & 0.3658      & 0.3041      & 0.3704 & \multicolumn{1}{c|}{0.2584} & N/A        & N/A        & N/A              & \multicolumn{1}{c|}{N/A}              & 0.4063        & 0.3746*       & 0.3976  & \multicolumn{1}{c|}{0.1751}  & 0.3898        & 0.2491   & 0.4111*  & \multicolumn{1}{c|}{0.2786}  & \textbf{0.4349}                           & \textbf{0.4040}     &5.79\%  & 7.85\%               \\\specialrule{0em}{1pt}{1pt}
\multicolumn{1}{c|}{\textbf{G-F}} & 0.3560      & 0.2631      & 0.2179 & \multicolumn{1}{c|}{0.1273} & N/A        & N/A        & N/A              & \multicolumn{1}{c|}{N/A}              & 0.3813        & 0.2495       & 0.4083*  & \multicolumn{1}{c|}{0.3292}  & 0.3792        & 0.1853   & 0.3927  & \multicolumn{1}{c|}{0.3513*}  & \textbf{0.4416}                           & \textbf{0.3941}      &8.16\%  & 12.18\%               \\\specialrule{0em}{1pt}{1pt}
\multicolumn{1}{c|}{\textbf{G-T}}  & 0.4335          & 0.3665          & 0.4539          & \multicolumn{1}{c|}{0.3366}          & N/A          & N/A          & N/A              & \multicolumn{1}{c|}{N/A}              & 0.4564          & 0.3479          & 0.4695*          & \multicolumn{1}{c|}{0.2983}          & 0.4327          & 0.2742          & 0.4633           & \multicolumn{1}{c|}{0.3784*}           & \textbf{0.5180}                                    & \textbf{0.4631}       &10.33\% & 22.38\%               \\ \specialrule{0.06em}{2pt}{2pt}

\multicolumn{21}{c}{\cellcolor[HTML]{E5E5E5}$\mathbf{\emph{K}_\emph{spt} = 3}$}                                                                                                                                                                                                                                                                                                                              \\\specialrule{0.06em}{2pt}{2pt}
\multicolumn{21}{c}{\textbf{Scenario 1: Single HG}}                                                                                                                                                                                                                                                                                                                                                                                              \\ \specialrule{0.06em}{2pt}{2pt}
\multicolumn{1}{c|}{\textbf{F}acebook}                                                    & 0.4863      & 0.3827      & 0.4285 & \multicolumn{1}{c|}{0.3230} & 0.5128     & 0.3971     & 0.5419*           & \multicolumn{1}{c|}{0.4967*}           & 0.5042        & 0.3423       & 0.5017  & \multicolumn{1}{c|}{0.4853}  & 0.4902        & 0.3125   & 0.5251  & \multicolumn{1}{c|}{0.4078}  & \textbf{0.5613}                           & \textbf{0.5329}      &3.58\%  & 7.29\%               \\ \specialrule{0em}{1pt}{1pt}
\multicolumn{1}{c|}{\textbf{T}witter}                                                     & 0.5258      & 0.4492      & 0.5044 & \multicolumn{1}{c|}{0.4275} & N/A        & N/A        & N/A              & \multicolumn{1}{c|}{N/A}              & 0.5347        & 0.3796       & 0.5642  & \multicolumn{1}{c|}{0.4408}  & 0.5229        & 0.3971   & 0.5862* & \multicolumn{1}{c|}{0.4667*} & \textbf{0.6018}                           & \textbf{0.5696}      & 2.66\%  & 22.05\%               \\ \specialrule{0em}{1pt}{1pt}
\multicolumn{1}{c|}{\textbf{G}oogle+}                                                     & 0.4231          & 0.3801          & 0.4387          & \multicolumn{1}{c|}{0.3192}          & 0.4767*          & 0.3864          & 0.4616  & \multicolumn{1}{c|}{0.4137*}  & 0.4428          & 0.3068          & 0.4532          & \multicolumn{1}{c|}{0.2985}          & 0.4681          & 0.3372          & 0.4597          & \multicolumn{1}{c|}{0.3958}          & \textbf{0.4946}                                    & \textbf{0.4712}       & 3.75\%  & 13.90\%               \\ \specialrule{0.06em}{2pt}{2pt}
\multicolumn{21}{c}{\textbf{Scenario 2: Multiple HGs \& single heterogeneity}}                                                                                                                                                                                                                                                                                                                                                                   \\ \specialrule{0.06em}{2pt}{2pt}
\multicolumn{1}{c|}{\textbf{G}oogle+}                                                     & 0.3975      & 0.2945      & 0.3674 & \multicolumn{1}{c|}{0.2814} & 0.4538*        & 0.3664        & 0.4427    &  \multicolumn{1}{c|}{0.3815*} & 0.4297        & 0.2816       & 0.4225 & \multicolumn{1}{c|}{0.2517} & 0.4194      & 0.2874     & 0.4418 & \multicolumn{1}{c|}{0.3557}     & \textbf{0.4894}       & \textbf{0.4213}      &7.84\%  & 10.43\%               \\ \specialrule{0.06em}{2pt}{2pt}
\multicolumn{21}{c}{\textbf{Scenario 3: Multiple HGs \& multiple heterogeneities}}                                                                                                                                                                                                                                                                                                                                                               \\ \specialrule{0.06em}{2pt}{2pt}
\multicolumn{1}{c|}{\textbf{F-T}} & 0.5044          & 0.3866          & 0.5114          & \multicolumn{1}{c|}{0.3762}          & N/A          & N/A          & N/A              & \multicolumn{1}{c|}{N/A}              & 0.5448          & 0.3850          & 0.5497*          & \multicolumn{1}{c|}{0.3819}          & 0.4877          & 0.3607          & 0.5298           & \multicolumn{1}{c|}{0.4284*}           & \textbf{0.5896}                           & \textbf{0.5456}      &7.26\%  & 27.36\%               \\\specialrule{0em}{1pt}{1pt}
\multicolumn{1}{c|}{\textbf{F-G}} & 0.4125          & 0.3361          & 0.4134          & \multicolumn{1}{c|}{0.2845}          & N/A             & N/A             & N/A              & \multicolumn{1}{c|}{N/A}              & 0.3936          & 0.3274          & 0.3906          & \multicolumn{1}{c|}{0.2480}          & 0.4065          & 0.2685          & 0.4693*          & \multicolumn{1}{c|}{0.3818*}          & \textbf{0.5034}                           & \textbf{0.4694}      &7.27\%  & 22.94\%               \\\specialrule{0em}{1pt}{1pt}
\multicolumn{1}{c|}{\textbf{T-F}} & 0.4165          & 0.3675          & 0.2612          & \multicolumn{1}{c|}{0.1715}          & N/A             & N/A             & N/A              & \multicolumn{1}{c|}{N/A}              & 0.4541          & 0.4117*          & 0.4871*          & \multicolumn{1}{c|}{0.3369}          & 0.4011          & 0.2255          & 0.4439          & \multicolumn{1}{c|}{0.3914}          & \textbf{0.4919}                           & \textbf{0.4611}      &0.99\%  & 12.00\%               \\\specialrule{0em}{1pt}{1pt}
\multicolumn{1}{c|}{\textbf{T-G}}  & 0.3847          & 0.3304          & 0.4139          & \multicolumn{1}{c|}{0.3236}          & N/A             & N/A             & N/A              & \multicolumn{1}{c|}{N/A}              & 0.4364          & 0.3918*          & 0.4043          & \multicolumn{1}{c|}{0.2287}          & 0.4469*          & 0.2668          & 0.4221          & \multicolumn{1}{c|}{0.3312}          & \textbf{0.5060}                           & \textbf{0.4769}      &13.22\% & 21.72\%               \\\specialrule{0em}{1pt}{1pt}
\multicolumn{1}{c|}{\textbf{G-F}} & 0.4420          & 0.3696          & 0.3232          & \multicolumn{1}{c|}{0.2135}          & N/A             & N/A             & N/A              & \multicolumn{1}{c|}{N/A}              & 0.3901          & 0.2655          & 0.4521*          & \multicolumn{1}{c|}{0.4180*}          & 0.3969          & 0.2664          & 0.4452          & \multicolumn{1}{c|}{0.3887}          & \textbf{0.4638}                           & \textbf{0.4259}      &2.59\%  & 1.89\%               \\\specialrule{0em}{1pt}{1pt}
\multicolumn{1}{c|}{\textbf{G-T}}  & 0.4863          & 0.3669          & 0.4377          & \multicolumn{1}{c|}{0.3744}          & N/A          & N/A          & N/A              & \multicolumn{1}{c|}{N/A}              & 0.5153          & 0.3887          & 0.5647*          & \multicolumn{1}{c|}{0.4084}          & 0.4848          & 0.3047          & 0.5504           & \multicolumn{1}{c|}{0.4294*}           & \textbf{0.5841}                           & \textbf{0.4956}      &3.44\%  & 15.42\%               \\ \specialrule{0.1em}{2pt}{2pt}
\end{tabular}}}
\footnotesize{\leftline{* Results of the best-performing baselines.}}\\
\footnotesize{\leftline{$^1$ Bold numbers are the results of the best-performing methods.}}
\footnotesize{\leftline{$^2$ Improvement of our proposed methods over the best-performing baseline.}}
\end{table*}

\subsection{Baselines}
Since there are no solutions specifically designed to solve the novel problem of few-shot semantic relation prediction, we selected 10 state-of-the-art models as baselines, which are relevant to our model and can be used to perform experiments with minor modifications. These baselines can be divided into five groups as follows:

\noindent \textbf{Semantic relation prediction models:} \emph{ProxEmbed} \cite{liu2017semantic} first samples multiple paths between two objects, then feeds these paths into LSTM to get the embedding for semantic relation prediction; \emph{IPE} \cite{liu2018interactive} constructs an interactive-path structure consisting of multiple paths between two objects, then adopts a Gated Recurrent Unit network to embed each interactive-path for semantic relation prediction.

\noindent \textbf{Homogeneous GNN:} \emph{GraphSAGE} \cite{hamilton2017inductive} is a graph neural network model that aggregates feature information of neighbors by different neural networks, such as LSTM; \emph{GAT} \cite{velivckovic2017graph} is a graph attention network model that aggregates neighbors’ feature information by self-attention neural network.
    
\noindent \textbf{Heterogeneous GNN:} \emph{HAN} \cite{wang2019heterogeneous} is a heterogeneous GNN that learns metapath-specific object embeddings from different metapath-based homogeneous graphs, and leverages the attention mechanism to combine them into one vector representation for each object; \emph{MAGNN} \cite{fu2020magnn} is a heterogeneous GNN that incorporates intermediate semantic objects along metapaths, and combines messages from multiple metapaths to learn object representations.
    
\noindent \textbf{Meta-learning based GNN:} \emph{Meta-GNN} \cite{zhou2019meta} is an MAML-based meta-learning framework that trains a GNN for few-shot node classification on homogeneous graphs; \emph{G-Meta} \cite{huang2020graph} uses local subgraphs with MAML to train a GNN for few-shot problems on homogeneous graphs.
    
\noindent \textbf{Subgraph-based GNN:} \emph{SEAL} \cite{zhang2018link} is a link prediction framework designed for homogeneous graphs, which simultaneously learns from local enclosing subgraphs, embeddings and attributes based on graph neural networks; \emph{SLiCE} \cite{wang2021self} is a graph neural network that aggregates information from higher-order relations in context subgraphs to learn contextual subgraph representations on heterogeneous graphs. Note that \emph{G-Meta} \cite{huang2020graph} is also a subgraph-based GNN model because it extracts local subgraphs for link prediction.

\subsection{Evaluation Metrics} 
We evaluate the performance of all the models from three perspectives, and the third one is newly proposed by us: (1) the \emph{prediction accuracy} to measure whether a model can accurately predict new semantic relations, with the support of few labeled data and the knowledge learnt from existing semantic relations; (2) the \emph{quality of object recommendation} to measure whether a model can effectively recommend a list of relevant objects for a query object with a new semantic relation; (3) the \emph{precision of recommending close objects} in the recommendation list, where “close objects” refer to objects that have any new semantic relation with the given query object. This perspective aims to measure whether a model is able to focus on objects that have close relationships with query objects, rather than investigating irrelevant objects.



Firstly, the metrics of existing methods only investigate the performance of a single semantic relation, while our work focuses on learning multiple new semantic relations across different graphs. Therefore, the prediction accuracy is assessed by two widely used multi-label classification metrics: \emph{accuracy} and \emph{Macro-F1 score} \cite{powers2011evaluation}. Secondly, for a query object with a new semantic relation, the object recommendation performance is evaluated by two metrics based on the top \emph{K} ranked objects: \emph{NDCG@K} and \emph{MAP@K}, which are commonly used in existing methods \cite{liu2017semantic,liu2018interactive}. Thirdly, the precision of recommending close objects (abbreviated as PRC) is defined as follows:
\begin{equation}
    \emph{PRC@K} = \frac{\emph{\#close\ objects}}{\emph{K}\times\emph{N}_{\emph{rel}}},
\end{equation}
where $\emph{N}_{\emph{rel}}$ is the number of new semantic relations to be predicted, \emph{\#close\ objects} is the total number of objects that have any new semantic relations with the query object in all $\emph{N}_{\emph{rel}}$ recommendation lists. This metric is necessary for the new problem of few-shot semantic relation prediction, because it can evaluate whether a model can effectively transfer knowledge from existing semantic relations, to identify close objects of new semantic relations with few labeled data.


\begin{figure*}[t]
\vspace{-0.1cm}  
\setlength{\abovecaptionskip}{-0.1cm}   
\setlength{\belowcaptionskip}{-1cm}   
\centering
\scalebox{0.40}{\includegraphics{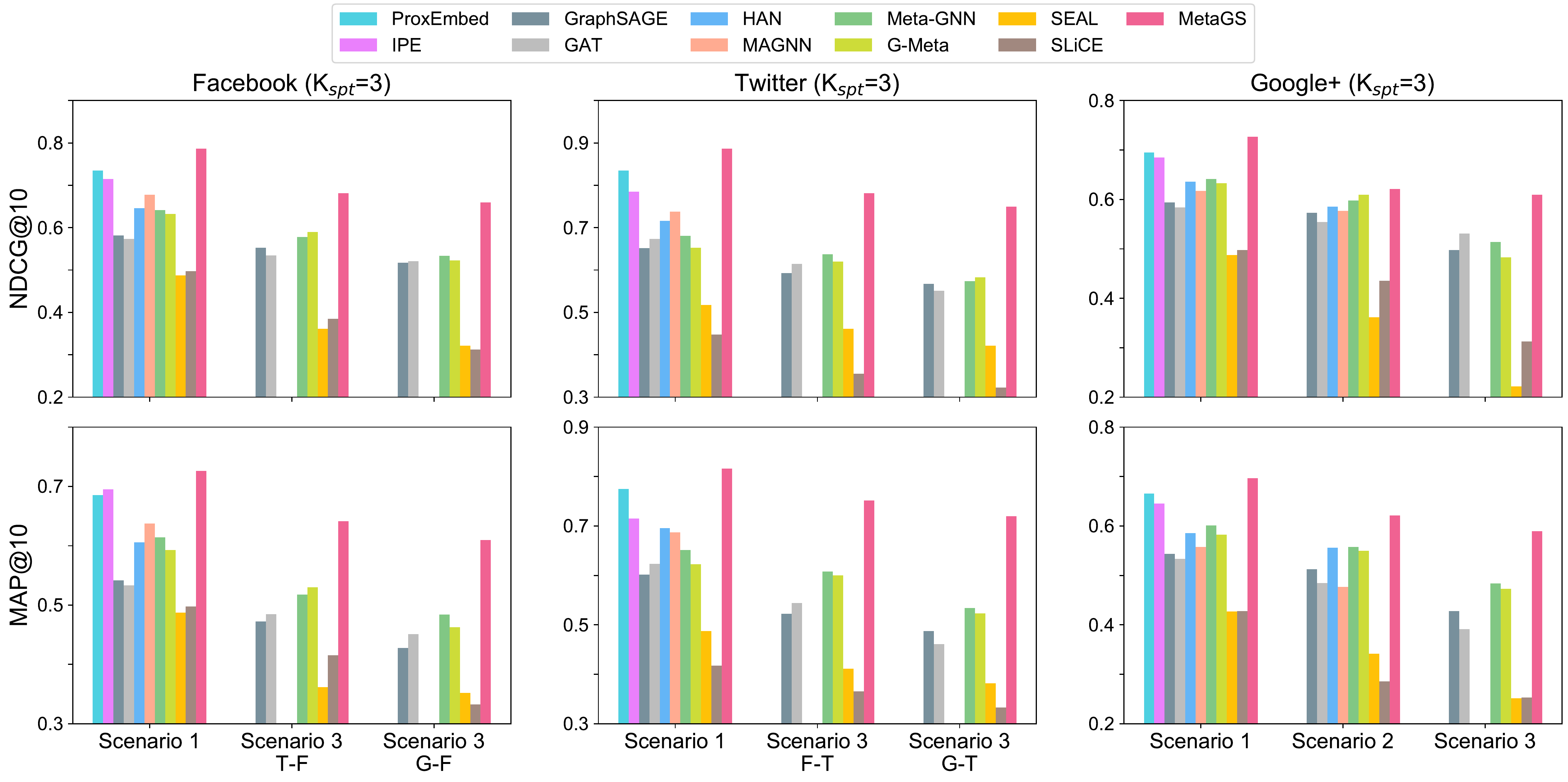}}
\caption{Comparison of object recommendation performance with baseline methods on three datasets.}
\label{map}
\end{figure*}

\subsection{Reproducibility Settings}
\subsubsection{Meta-learning Settings} 
We adopt three different meta-learning settings to investigate the performance on the three scenarios of few-shot semantic relation prediction as follows:

\noindent\textbf{Scenario 1: Single HG.} In this scenario, few-shot semantic relation prediction is performed on a single graph. For each dataset, we first select multiple graphs, and then average the performance on these graphs to obtain results for the dataset. Specifically, for Facebook dataset, we perform experiments on its all 10 graphs; for Twitter and Google+ datasets, we randomly select 50 graphs in each dataset for experiments. In each graph, we randomly select 5 semantic relations for meta-training, 3 other semantic relations for meta-validation, and the rest semantic relations for meta-testing. 

\noindent\textbf{Scenario 2: Multiple HGs \& single heterogeneity.} In this scenario, few-shot semantic relation prediction is performed on multiple graphs from a single dataset. We only perform experiments on the Google+ dataset because all graphs in the dataset consist of fixed object types. Graphs in any of the other two datasets (Facebook or Twitter) have different heterogeneities (i.e., different object types). Specifically, in Google+ dataset, we select all semantic relations in 4 HGs (sum to about 16 different semantic relations) randomly for meta-training, other 2 HGs (sum to 8 relations) for meta-validation, and 4 HGs (sum to 16 relations) from the remaining graphs for meta-testing.

\noindent\textbf{Scenario 3: Multiple HGs \& multiple heterogeneities.} In this scenario, we perform meta-training and meta-validation on one of the three datasets, then use another dataset for meta-testing. Firstly, we randomly select one dataset, then select all semantic relations in 4 HGs (sum to about 40 different semantic relations) for meta-training, and other 2 HGs (sum to 20 relations) for meta-validation. Secondly, we randomly choose another dataset and select 4 HGs (sum to 40 relations) of the dataset randomly for meta-testing. 

For each meta-learning setting, we perform 10 independent runs and report the average results. In each run, the task setting is the same for all the three scenarios. Specifically, for a task $\mathcal{T}$ = $(\mathcal{T}^\emph{spt}, \mathcal{T}^\emph{qry})$, the number of samples for one semantic relation in support set $\mathcal{T}^\emph{spt}$, i.e., $\emph{K}_\emph{spt}$, is set to \{1, 3, 5\}; and the number of samples for one semantic relation in query set $\mathcal{T}^\emph{qry}$, i.e., $\emph{K}_\emph{qry}$, is set to 10.

\subsubsection{Parameter Settings} 
Among all the ten baselines, except for ProxEmbed and IPE, the other eight GNN-based methods are not designed for semantic relation prediction. Therefore, we modify these GNN-based methods to achieve few-shot semantic relation prediction. The modification is two-fold. On the one hand, to realize semantic relation prediction, subgraph based GNNs (SEAL, SLiCE and G-Meta) encode the subgraph between a pair of objects to predict semantic relations of the two objects. The rest GNNs encode all objects in the HG and concatenate embeddings of two objects for semantic relation prediction. On the other hand, because meta-learning based GNNs (Meta-GNN and G-Meta) already have a meta-learning module, we integrate GPN \cite{ding2020graph} with the rest GNN-based methods to achieve meta-learning. For fair comparisons, all the model parameters including hyper-parameters of both baseline methods and our method are well tuned in the same way through meta-validation. Specifically, for the parameters in each baseline, we first initialize them with the values reported in the original paper and then carefully tune them on our datasets for best performance. The embedding dimensions for objects and semantic relations are set to 64 and 256 respectively. Other important parameters for each baseline will be specified during the experimental result analysis in Section \ref{comp_baseline}. 

For our proposed MetaGS, in the subgraph extraction module, we adopt node2vec \cite{grover2016node2vec} to generate $\mathbf{e}$ in E.q. (\ref{type_sim}). $\emph{K}_\emph{path}$ is set to 500, $\emph{N}_\emph{subg}$ is set to 100, and $\emph{N}_\emph{type}$ is set to [2, 6]. In OVGC module, $f_\emph{agg}(\cdot)$ is set to $\emph{sum}(\cdot)$ and neighbor hop $h$ is set to 2. In GVGC module, we use GraphSAGE \cite{hamilton2017inductive} as the base model, the activation $\sigma(\cdot)$ is set to $\emph{relu}(\cdot)$, $\emph{gv-pool}(\cdot)$ is set to $\emph{mean-pool}(\cdot)$, and the number of layers is set to 2. In the hyper-prototypical network module, $\theta_\emph{ho}$ and $\theta_\emph{he}$ are set to 0.75 and 0.15 respectively, $l$ is set to 2, and $\emph{hy-pool}(\cdot)$ is set to $\emph{mean-pool}(\cdot)$. The study of these parameters can be seen in appendix.





\subsection{Performance Comparison with Baselines}
\label{comp_baseline}
\subsubsection{Comparisons w.r.t. Prediction Accuracy}
Since ProxEmbed and IPE are designed to learn a particular semantic relation in a particular HG, they cannot be used to learn multiple new semantic relations across different HGs. Therefore, the prediction accuracy of our proposed MetaGS is compared to the rest eight baselines, and the results are shown in Table \ref{exp_res}. “F-T” stands for training and validating on Facebook, testing on Twitter. For the fair comparison, we carefully tuned the parameters of each baseline through meta-validation. In GraphSAGE, the number of layers is set to 1 and the aggregator function is set to GCN. In HAN and GAT, the number of layers is set to 1 and the the number of attention heads is set to 4. In MAGNN, the number of attention heads is set to 8 and the relational rotation encoder is used to encode objects. In G-Meta, 10 gradient update steps are used in meta-training and 20 gradient update steps are used in meta-testing. In SEAL, the hop number is set to 2. In SLiCE, the context subgraphs are generated by the shortest path strategy.

From Table \ref{exp_res} we can see: (1) MetaGS shows significant improvement over the best-performing baselines (with results marked by *) by an average of 5.68\% in accuracy and an average of 17.03\% in F1$\emph{-}$score. The reason is that MetaGS adopts the two-view GNN to learn the generalized knowledge of both various subgraph structures and multiple heterogeneities simultaneously, and thus outperforming baselines that only utilize information of particular graph structures or a fixed heterogeneity; (2) With the decrease of the number of labeled data (i.e., $\emph{K}_\emph{spt}$), MetaGS achieves a larger improvement over the best-performing baselines. Specifically, when $\emph{K}_\emph{spt}$=$3$, the average improvements over the best-performing are 5.26\% and 15.50\% in accuracy and F1$\emph{-}$score, respectively. The values rise to 6.10\% and 18.55\% when $\emph{K}_\emph{spt}$=$1$. This is because MetaGS adopts the hyper-prototypical network to effectively learn new semantic relations with few labeled samples, which extracts important information while filtering out worthless information. Thus, MetaGS can achieve satisfactory performance when labeled data is limited; (3) MetaGS achieves the best performance in all the three scenarios. Specifically, in Scenario 1 (Single HG), MetaGS improves the best-performing baselines by an average of 4.67\% in accuracy and an average of 17.07\% in F1$\emph{-}$score; in Scenario 2 (Multiple HGs \& single heterogeneity), the average improvements are 5.19\% in accuracy and 15.31\% in F1$\emph{-}$score; in Scenario 3 (Multiple HGs \& multiple heterogeneities), the average improvements are 6.26\% in accuracy and 17.29\% in F1$\emph{-}$score. In Scenarios 1 and 2, subgraph-based GNNs (SEAL and SLiCE) and heterogeneous GNNs (HAN and MAGNN) perform better than other baselines in most cases, demonstrating that learning local heterogeneous information and graph structure between objects is important for semantic relation prediction in these two scenarios. In Scenario 3, Meta-GNN and G-Meta surpass other baselines due to specifically designed meta-learning modules. In contrast, our proposed MetaGS employs not only the two-view GNN to learn local heterogeneous information and graph structure, but also a meta-learning based hyper-prototypical network to transfer knowledge across semantic relations. Therefore, MetaGS can effectively deal with all the three scenarios.



\subsubsection{Comparisons w.r.t. the Performance of Object Recommendation} Existing semantic relation prediction methods (ProxEmbed and IPE) target a single HG and thus cannot deal with Scenario 2 (Multiple HGs \& single heterogeneity) and Scenario 3 (Multiple HGs \& multiple heterogeneities). Since heterogeneous GNNs (HAN and MAGNN) target a single heterogeneity, they cannot deal with Scenario 3. Our MetaGS consistently outperforms all the baselines in all $\emph{K}_\emph{spt}$ settings, i.e., $\{1,3,5\}$, and demonstrates similar improvements over the best-performing baselines. Due to the limited space, we only report the results of $\emph{K}_\emph{spt}$=$3$. In Fig. \ref{map}, it is obvious that our proposed MetaGS outperforms all compared baseline methods with a clear margin w.r.t. NDCG and MAP on the three datasets. Specifically, on Facebook, MetaGS improves the best-performing baselines in all scenarios by an average of 15.34\% for NDCG@10, and by an average of 17.18\% for MAP@10. The values are 19.09\% and 21.25\% on Twitter, and are 7.04\% and 12.65\% on Google+. This is due to the fact that these baselines do not focus on capturing heterogeneous information and graph structure information to characterise features of a specific semantic relation. Therefore, these baselines may not correctly recommend objects that have the desired new semantic relation to the query object. In contrast, in our proposed MetaGS, the graph structure between objects is fully analysed to differentiate all possible simple and complex relationships between objects. Hence, MetaGS can effectively characterise features of new semantic relations with few labeled data, and thus outperforms baselines in object recommendation.

\begin{figure}[t]
\setlength{\abovecaptionskip}{-0.05cm}   
\setlength{\belowcaptionskip}{-1cm}   
\centering
\scalebox{0.36}{\includegraphics{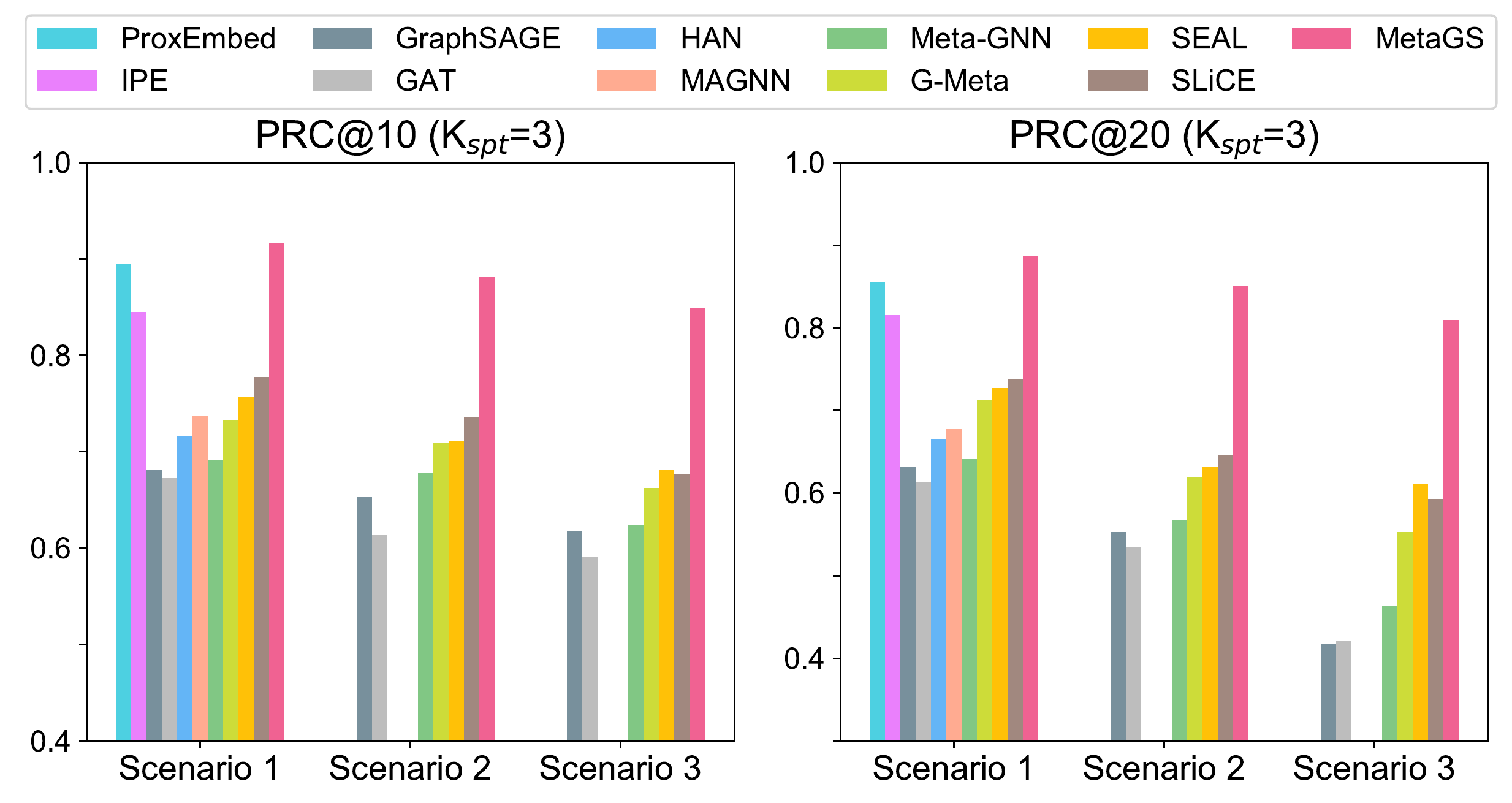}}
\caption{The precision of recommending close objects (PRC) in recommendation lists on Google+ dataset.}
\label{rc}
\end{figure}

\subsubsection{Comparisons w.r.t. the Precision of Recommending Close Objects}
Fig. \ref{rc} reports the precision of recommending close objects (PRC) in the recommendation lists. MetaGS achieves the best performance in all datasets, and we only report the performance on the Google+ dataset due to space limitations. In Fig. \ref{rc}, MetaGS has the best performance in all the three scenarios. Specifically, MetaGS improves the best-performing baselines in all scenarios by an average of 15.61\% for PRC@10, and by an average of 22.63\% for PRC@20. The reasons are two-fold. On the one hand, GNN based baselines focus on encoding objects by aggregating information from their neighbors while ignoring mining the relationships between objects for semantic relation prediction. Therefore, these GNN based baselines may recommend irrelevant objects that only have similar neighbors or features to the query object. On the other hand, existing semantic relation prediction models focus on learning a single semantic relation in a particular graph, and thus cannot leverage information from multiple existing semantic relations across multiple graphs, to specify objects with close relationships to the query object. In contrast, MetaGS can effectively learn and generalize the information between close objects from multiple existing semantic relations, and thus can easily identify close objects for a new semantic relation.

It is worth noting that subgraph-based GNN models (SEAL, SLiCE and G-Meta) outperform better than other GNN baselines. The reason is that these subgraph-based GNN models mine the graph structure between objects for semantic relation prediction, and thus can filter out irrelevant objects that have no connections to the query object. However, these models directly encode the whole graph structure between objects and are unable to generalize the heterogeneous information across multiple graphs. Thus, these methods cannot perform better than our MetaGS, which decomposes the graph structure to learn generalized knowledge across different graphs.

\begin{table}[]
\centering
\setlength{\abovecaptionskip}{-0.05cm}   
\setlength{\belowcaptionskip}{-1cm}   
\caption{Comparison of MetaGS with its variants on three datasets.}
\label{exp_abl}
\resizebox{88mm}{54mm}
{\setlength{\tabcolsep}{1mm}{
\begin{tabular}{ccccccccccc}
\specialrule{0.06em}{2pt}{2pt}
\multicolumn{1}{c|}{}                                                            & \multicolumn{2}{c}{\textbf{MetaGS}} & \multicolumn{2}{c}{\textbf{MetaGS-2Hop}} & \multicolumn{2}{c}{\textbf{MetaGS-OV}} & \multicolumn{2}{c}{\textbf{MetaGS-GV}} & \multicolumn{2}{c}{\textbf{MetaGS-Hyper}} \\ \specialrule{0.06em}{2pt}{2pt}
\multicolumn{1}{c|}{\multirow{-2}{*}{\textbf{}}}                                 & \textbf{Acc}          & \textbf{F1}          & \textbf{Acc}          & \textbf{F1}         & \textbf{Acc}          & \textbf{F1}           & \textbf{Acc}          & \textbf{F1}          & \textbf{Acc}          & \textbf{F1}         \\ \specialrule{0.06em}{2pt}{2pt}
\multicolumn{11}{c}{\cellcolor[HTML]{E5E5E5}$\mathbf{\emph{K}_\emph{spt} = 1}$}                                                                                                                                                                         \\ \specialrule{0.06em}{2pt}{2pt}
\multicolumn{11}{c}{ \textbf{Scenario 1: Single HG}}                                                                                                                                                                        \\ \specialrule{0.06em}{2pt}{2pt}
\multicolumn{1}{c|}{\textbf{F}acebook}                                                    & \textbf{0.4975}      & \textbf{0.4553}      & 0.4713     & 0.4329     & 0.4626        & 0.4170       & 0.4473       & 0.3860      & 0.4235      & 0.3977     \\ \specialrule{0em}{1pt}{1pt}
\multicolumn{1}{c|}{\textbf{T}witter}                                                    & \textbf{0.5295}      & \textbf{0.4951}      & 0.4829     & 0.4722     & 0.4796        & 0.4629       & 0.4892       & 0.4645      & 0.4937      & 0.4127     \\ \specialrule{0em}{1pt}{1pt}
\multicolumn{1}{c|}{\textbf{G}oogle+}                                                   & \textbf{0.4786}      & \textbf{0.4141}      & 0.4437     & 0.3704     & 0.4512        & 0.3865       & 0.4413       & 0.3981      & 0.4331      & 0.3680     \\ \specialrule{0.06em}{2pt}{2pt}
\multicolumn{11}{c}{\textbf{Scenario 2: Multiple HGs \& single heterogeneity}}                                                                                                                                                                     \\ \specialrule{0.06em}{2pt}{2pt}
\multicolumn{1}{c|}{\textbf{G}oogle+}                                                     & \textbf{0.4435}      & \textbf{0.4044}      & 0.4125     & 0.3921     & 0.4251        & 0.3816       & 0.4028       & 0.3715      & 0.3914      & 0.3658     \\ \specialrule{0.06em}{2pt}{2pt}
\multicolumn{11}{c}{\textbf{Scenario 3: Multiple HGs \& multiple heterogeneities} }                                                                                                                                                                 \\ \specialrule{0.06em}{2pt}{2pt}
\multicolumn{1}{c|}{\textbf{F-T}} & \textbf{0.5483}      & \textbf{0.5067}      & 0.4913     & 0.4840     & 0.4978        & 0.4862       & 0.4955       & 0.4685      & 0.5193      & 0.4593     \\ \specialrule{0em}{1pt}{1pt}
\multicolumn{1}{c|}{\textbf{F-G}} & \textbf{0.4451}      & \textbf{0.4147}      & 0.3874     & 0.3577     & 0.4285        & 0.3655       & 0.3906       & 0.3680      & 0.3975      & 0.3811     \\ \specialrule{0em}{1pt}{1pt}
\multicolumn{1}{c|}{\textbf{T-F}} & \textbf{0.4673}      & \textbf{0.4237}      & 0.4468     & 0.3968     & 0.4059        & 0.3883       & 0.4607       & 0.4168      & 0.4477      & 0.3921     \\ \specialrule{0em}{1pt}{1pt}
\multicolumn{1}{c|}{\textbf{T-G}}  & \textbf{0.4349}      & \textbf{0.4040}      & 0.4177     & 0.3854     & 0.4216        & 0.3921       & 0.4127       & 0.3751      & 0.4087      & 0.3715     \\ \specialrule{0em}{1pt}{1pt}
\multicolumn{1}{c|}{\textbf{G-F}} & \textbf{0.4416}      & \textbf{0.3941}      & 0.3925     & 0.3350     & 0.4195        & 0.3855       & 0.4272       & 0.3897      & 0.4125      & 0.3629     \\ \specialrule{0em}{1pt}{1pt}
\multicolumn{1}{c|}{\textbf{G-T}}  & \textbf{0.5180}      & \textbf{0.4631}      & 0.4959     & 0.4095     & 0.5076        & 0.4325      & 0.4895       & 0.3981      & 0.4741      & 0.3862     \\ \specialrule{0.06em}{2pt}{2pt}
\multicolumn{11}{c}{\cellcolor[HTML]{E5E5E5}$\mathbf{\emph{K}_\emph{spt} = 3}$}                                                                                                                                                                         \\ \specialrule{0.06em}{2pt}{2pt}
\multicolumn{11}{c}{\textbf{Scenario 1: Single HG}}                                                                                                                                                                        \\  \specialrule{0.06em}{2pt}{2pt}
\multicolumn{1}{c|}{\textbf{F}acebook}                                                    & \textbf{0.5613}      & \textbf{0.5329}      & 0.5534     & 0.5093     & 0.4939        & 0.4561       & 0.5435       & 0.5278      & 0.5535      & 0.5477     \\ \specialrule{0em}{1pt}{1pt}
\multicolumn{1}{c|}{\textbf{T}witter}                                                    & \textbf{0.6018}      & \textbf{0.5696}      & 0.5968     & 0.5574     & 0.5912        & 0.5244       & 0.5887       & 0.5465      & 0.5627      & 0.5332     \\ \specialrule{0em}{1pt}{1pt}
\multicolumn{1}{c|}{\textbf{G}oogle+}                                                   & \textbf{0.4946}      & \textbf{0.4712}      & 0.4521     & 0.4141     & 0.4407        & 0.4057       & 0.4573       & 0.4078      & 0.4613      & 0.4348     \\ \specialrule{0.06em}{2pt}{2pt}
\multicolumn{11}{c}{\textbf{Scenario 2: Multiple HGs \& single heterogeneity}}                                                                                                                                                                     \\ \specialrule{0.06em}{2pt}{2pt}
\multicolumn{1}{c|}{\textbf{G}oogle+}                                                     & \textbf{0.4894}      & \textbf{0.4213}      & 0.4653     & 0.4013     & 0.4436        & 0.4172       & 0.4486       & 0.4087      & 0.4214      & 0.3968     \\ \specialrule{0.06em}{2pt}{2pt}
\multicolumn{11}{c}{\textbf{Scenario 3: Multiple HGs \& multiple heterogeneities}}                                                                                                                                                                 \\ \specialrule{0.06em}{2pt}{2pt}
\multicolumn{1}{c|}{\textbf{F-T}} & \textbf{0.5896}      & \textbf{0.5456}      & 0.5525     & 0.5132     & 0.5427        & 0.5025       & 0.5569       & 0.4987      & 0.5304      & 0.4965     \\ \specialrule{0em}{1pt}{1pt}
\multicolumn{1}{c|}{\textbf{F-G}} & \textbf{0.5034}      & \textbf{0.4694}      & 0.4839     & 0.4439     & 0.4448        & 0.4108       & 0.4877       & 0.4365      & 0.4940      & 0.4053     \\ \specialrule{0em}{1pt}{1pt}
\multicolumn{1}{c|}{\textbf{T-F}} & \textbf{0.4919}      & \textbf{0.4611}      & 0.4615     & 0.4196     & 0.4554        & 0.4029       & 0.4773       & 0.4310      & 0.4604      & 0.4132     \\ \specialrule{0em}{1pt}{1pt}
\multicolumn{1}{c|}{\textbf{T-G}}  & \textbf{0.5060}      & \textbf{0.4769}      & 0.4838     & 0.4573     & 0.4495        & 0.4044       & 0.4867       & 0.4602      & 0.4803      & 0.4486     \\ \specialrule{0em}{1pt}{1pt}
\multicolumn{1}{c|}{\textbf{G-F}} & \textbf{0.4638}      & \textbf{0.4259}      & 0.4174     & 0.3811     & 0.4233        & 0.3908       & 0.4398       & 0.3959      & 0.4319      & 0.3815     \\ \specialrule{0em}{1pt}{1pt}
\multicolumn{1}{c|}{\textbf{G-T}}  & \textbf{0.5841}      & \textbf{0.4956}      & 0.5694     & 0.4746     & 0.5669        & 0.4705       & 0.5458       & 0.4437      & 0.5289      & 0.4788     \\ \specialrule{0.06em}{2pt}{2pt}

\end{tabular}}}
\end{table}

\subsection{Ablation Study} 
\subsubsection{Settings}
We conduct an ablation study to analyse the rationality and the effectiveness of the designed components in our model. Specifically, we compare the semantic relation prediction performance of the original MetaGS with that of its four variants:

\noindent\textbf{MetaGS-2Hop}: The variant that removes the graph structure extraction module (cf. Section \ref{sub_extract}) by extracting 2 hop neighbors around objects $u$ and $v$ to generate the graph structure between them (i.e., $\mathcal{S}_{u,v}$).

\noindent \textbf{MetaGS-OV}: The variant that removes the object-view graph convolution (OVGC) module (cf. Section \ref{ovgc}). This variant does not model heterogeneous information of all objects in the subgraph, and only uses graph-view graph convolution (GVGC) module (cf. Section \ref{gvgc}) to encode each subgraph (i.e., $\mathbf{z}_{u, v}$ = $\mathbf{z}^{gv}_{u, v}$).

\noindent \textbf{MetaGS-GV}: The variant that removes the GVGC module. This variant does not leverage the structure information of the subgraph, and only uses OVGC module to encode each subgraph (i.e., $\mathbf{z}_{u, v}$ = $\mathbf{z}^{ov}_{u, v}$).

\noindent \textbf{MetaGS-Hyper}: The variant that removes the hyper-prototypical network module (cf. Section \ref{hyper_net}). This variant encodes a sample by taking the average of its all subgraphs' embeddings (i.e., $\mathbf{z}_{u, v}$), and learns semantic relation representation by taking the average of embedded samples that belong to a semantic relation.



\subsubsection{Observations}
The results are reported in Table \ref{exp_abl} and MetaGS significantly outperforms each of its four variants. From this table, we have the following observations:

\noindent \textbf{Observation 1}: \emph{Graph structure extraction module can obviously boost the prediction accuracy} by selecting useful paths that connect objects to ensure all possible relations between objects can be mined. Specifically, by comparing the values in the first four columns in Table \ref{exp_abl}, we can see that the graph structure extraction module leads to an average improvement of 6.76\% in accuracy and an average improvement of 7.80\% in F1-score. 

\noindent \textbf{Observation 2}: \emph{OVGC module and GVGC module can greatly improve the performance} by capturing generalized heterogeneous information and graph structure information across multiple HGs, respectively. From Table \ref{exp_abl}, we can see that the OVGC module leads to an average improvement of 8.05\% in accuracy and an average improvement of 9.03\% in F1-score; the GVGC module leads to an average improvement of 6.50\% in accuracy and an average improvement of 7.79\% in F1-score.

\noindent \textbf{Observation 3}: \emph{The hyper-prototypical network clearly benefits the few-shot semantic relation prediction task} by utilising a meta-learning based framework that can learn how to non-linearly aggregate subgraphs for semantic relation representation from existing semantic relations, and then transfer the learned knowledge to adapt to new semantic relations with few labeled data. From Table \ref{exp_abl}, we can see that the hyper-prototypical network leads to an average improvement of 8.40\% in accuracy and the average improvement of 9.97 \% in F1-score.

\section{Conclusion and Future Work}
In this paper, we have proposed a novel problem of few-shot semantic relation prediction across heterogeneous graphs, and provide a solution, \textbf{Meta}-learning based \textbf{G}raph neural network for \textbf{S}emantic relation prediction, named MetaGS. MetaGS is also a general framework that works for all three scenarios of few-shot semantic relation prediction. The two-view graph neural network and hyper-prototypical network in our MetaGS framework effectively transfer knowledge of existing semantic relations to learn new semantic relations with few labeled data. We conduct extensive experiments to demonstrate the superior performance of our proposed MetaGS. In the future, we plan to extend our approach to dynamic heterogeneous graphs where objects and links change over time.

\bibliographystyle{IEEEtran}
\bibliography{reference}

\end{document}